\documentclass[acmtog, nonacm]{acmart}

\renewcommand\footnotetextcopyrightpermission[1]{} 
\makeatletter
\let\@authorsaddresses\@empty
\def\runningfoot{\def\@runningfoot{}}
\def\firstfoot{\def\@firstfoot{}}
\makeatother

\usepackage[ruled,vlined]{algorithm2e}
\usepackage{subcaption}
\usepackage{float}
\usepackage{rotating}
\usepackage{appendix}
\usepackage[export]{adjustbox}
\usepackage{multirow}

\AtBeginDocument{%
  \providecommand\BibTeX{{%
    \normalfont B\kern-0.5em{\scshape i\kern-0.25em b}\kern-0.8em\TeX}}}

\begin{document}

\title{Latent-Insensitive Autoencoders for Anomaly Detection}


\author{Muhammad S. Battikh}
\email{mohammud.saeed.batekh@gmail.com}
\affiliation{%
  \institution{System and  Computer Engineering Department, Al-Azhar University in Cairo}
  \city{Cairo}
  \state{Cairo}
  \country{Egypt}
}

\author{Artem A. Lenskiy}
\email{Artem.Lenskiy@anu.edu.au}
\affiliation{%
  \institution{School of Computing, The Australian National University}
  \city{Canberra}
  \state{ACT}
  \country{Australia}
}


\begin{abstract}
Reconstruction-based approaches to anomaly detection tend to fall short when applied to complex datasets with target classes that possess high inter-class variance \cite{gong2019memorizing, perera2019learning}. Similar to the idea of self-taught learning \cite{raina2007self} used in transfer learning, many domains are rich with \textit{similar} unlabeled datasets that could be leveraged as a proxy for out-of-distribution samples. In this paper we introduce Latent-Insensitive autoencoder (LIS-AE) where unlabeled data from a similar domain is utilized as negative examples to shape the latent layer (bottleneck) of a regular autoencoder such that it is only capable of reconstructing one task. 
We provide theoretical justification for the proposed training process and loss functions along with an extensive ablation study highlighting important aspects of our model. We test our model in multiple anomaly detection settings presenting quantitative and qualitative analysis showcasing the significant performance improvement of our model for anomaly detection tasks.

\end{abstract}



\keywords{Anomaly Detection, autoencoders, One-Class Classification, Principal Components Analysis, Self-Taught Learning, Negative Examples.}




\maketitle

\section{Introduction}
Anomaly detection is a classical machine learning field which is concerned with the identification of in-distribution and out-of-distribution samples that finds applications in numerous fields \cite{zhou2017anomaly, ahmad2017streaming}. Unlike traditional multi-label classification where the goal is to find decision boundaries between classes present in a given dataset, the goal of anomaly detection is to find one-versus-all boundaries for classes that are not in the dataset which is significantly more challenging compared to standard classification. Autoencoders \cite{bengio2007greedy} have been used extensively for anomaly detection \cite{zhou2017anomaly,zimek2012survey,chalapathy2017robust} under the assumption that reconstruction error incurred by anomalies is higher than that of normal samples \cite{hasan2016learning, zong2018deep}. However, it has been observed that this assumption might not hold as standard autoencoders might generalize so well even for anomalies\cite{gong2019memorizing, zong2018deep}. In practice, this issue becomes more relevant in two important settings, namely, when the normal data is relatively complex it requires high latent dimension for \textit{good} reconstitution, and when anomalies share similar compositional features and are from a close domain to the normal data \cite{chandola2009anomaly}.\\
To mitigate these issues we present Latent-Insensitive autoencoder (LIS-AE), a new class of autoencoders where the training process is carried out in two phases. In the first phase, the model simply reconstructs the input as a standard autoencoder, in the second phase the entire model except the latent layer is "\textit{frozen}". We then train the model in such a way that forces the latent layer to only keep reconstructing the target task. We use the concept of a negative dataset from one-class classification \cite{weston2006inference} whereby an auxiliary dataset of non-examples from similar domains is used as a proxy for out-of-distribution samples. \textit{We change the training objective such that the autoencoder keeps its low reconstruction error for the target dataset while pushing the error of the negative dataset to exceed certain value.} In some cases, minimizing and maximizing the reconstruction loss at the same time becomes contradictory, especially for negative classes that are very similar to the target class. To resolves this issue we introduce another variant with modified first phase loss that ensures that the input of the latent layer is linearly-separable for positive and negative examples during the second phase. This linearly-separable variant (LinSep-LIS-AE) almost always performs better than directly using LIS-AE. Details of architecture, training process, theoretical analysis, and experiments are discussed in detail in the following sections.

\begin{figure}
 \begin{subfigure}{\linewidth}
    \centering
    \includegraphics[width=0.8\linewidth, clip, trim={8cm 9cm 8cm 8cm}]{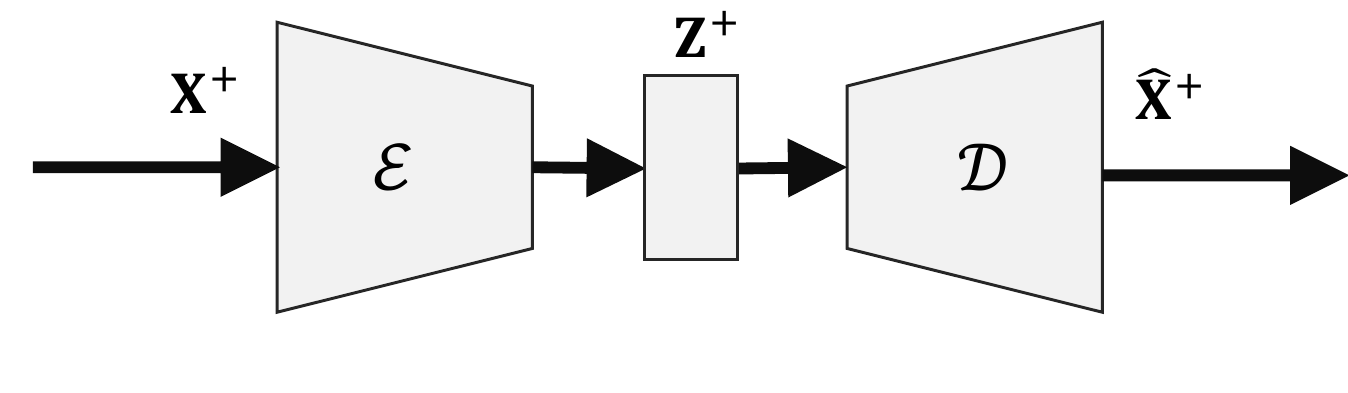} 
    \caption{The first diagram shows feature extraction phase.} 
    \label{fig1:a} 
    \vspace{0ex}
  \end{subfigure}
  
  \begin{subfigure}{\linewidth}
    \centering
    \includegraphics[width=0.8\linewidth, clip, trim={8cm 9cm 8cm 6cm}]{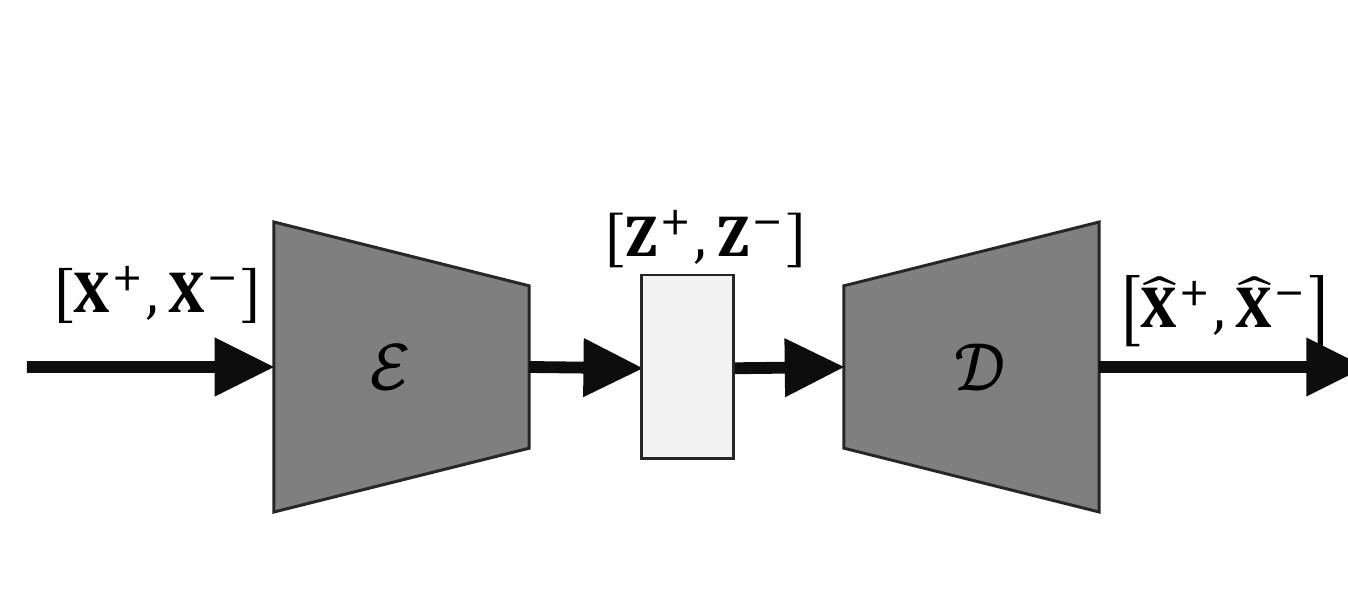} 
    \caption{The second phase starts by freezing the model except latent layer. Negative examples $\mathbf{x^-}$ are used to fine-tune the latent layer to be only responsive to $\mathbf{x^+}$.} 
    \label{fig1:b} 
    \vspace{0ex}
  \end{subfigure}
  \caption{The two phases of training.}
    \label{fig:fig1}
\end{figure}

\section{Related Work}

Many reconstruction-based anomaly detection approaches have been proposed starting with classical methods such as PCA \cite{pearson1901liii}. Robust-PCA mitigates the issue of outlier sensitivity in PCA by decomposing the data matrix into a sum of two low-rank and sparse matrices using nuclear norm and $L_1$ norms as convex relaxation for the objective loss \cite{candes2011robust}. autoencoders address the issue of PCA only considering linear relations in feature-space by introducing nonlinearities benefiting from multiple layers of representations \cite{bourlard1988auto}. We elaborate further on the shortcomings of PCA and autoencoders in the theoretical section and use that to motivate our approach. 
 
Other methods try to improve on base autoencoders by endowing the latent code with particular properties. In the case of VAE \cite{an2015variational}, it does so by having the latent code to follow a prior distribution (usually normal) which also allows sampling from the decoder. However, in the context of anomaly detection, it introduces scaling issues since minimizing KL-Divergence for high latent dimensions required for complex tasks is quite challenging. 
Another approach is Replicator Neural Networks (RepNN) \cite{hawkins2002outlier} which is an autoencoder with a staircase activation function positioned on the output of the bottleneck layer (Latent Layer). This is mainly used in order to quantize the latent code into a number of discrete values which also aids in forming clusters \cite{williams2002comparative}. Unfortunately, a discrete staircase function is non-differentiable which prevents learning via backpropagation. Instead, a differentiable approximation involving the sum of $N$ hyperbolic tangent functions \texttt{tanh} was introduced in place of the otherwise, non-differentiable discrete staircase function. However, as discussed in \cite{toth2004replicator}, despite the theoretical appeal for having a quantized latent code via smooth approximation, in practice, having such activation function makes it significantly difficult for the gradient signal to flow. We also note that increasing the number of levels using the aforementioned $N$, \texttt{tanh} sum approximation presents a significant overhead during training and testing since $N$ activation functions have to be computed for each batch, moreover, it suffered from scaling issues similar to that of VAE.\\
Another approach is \textit{memorizing normality} of a given dataset using Memory-augmented autoencoder \cite{gong2019memorizing}. This approach limits the effective space of possible latent codes by constructing a memory module that takes in the output of the encoder as an address and passes to the decoder the most relevant memory items from a stored reservoir of prototypical patterns that have been learned during training.\\
Other non-reconstruction-based approaches include One-Class classification which is tightly connected to anomaly detection in the sense that both problems are concerned with finding one-versus-all boundaries. One-Class SVM is a variation of the classical SVM algorithm \cite{cortes1995support} where the objective is to find a hyper-plane that best separates samples from outliers \cite{scholkopf2001estimating}. Support Vector Data Description (SVDD)\cite{tax2004support} tries to find a circumscribing hyper-sphere that contains all samples while having optimal margin for outliers. It is worth noting that for kernels where $k(x,x)=1$ such as  RBF and Laplacian, OC-SVM and OC-SVDD learn identical decision functions \cite{lampert2009kernel}. To address the lack of representation learning and bad computational scalability of OC-SVM and OC-SVDD, Deep SVDD (OC-DSVDD) employs a deep neural network that learns useful representation while mapping outputs to a hypersphere of minimum volume \cite{ruff2018deep}. However, due to its sole reliance  on optimizing for minimum volume, this approach is prone to hyper-sphere collapse which leads to finding uninformative features \cite{perera2019learning}.

Other approaches have been proposed where an auxiliary datset of non-examples (negative dataset) is drawn from similar domains as a proxy for the otherwise intractable complement for the target class. In \cite{weston2006inference}, a collection called the "\textit{Universum}", allows learning useful representation to the domain of the problem via maximizing the number of contradictions on an equivalence class.
Similar to OC-DSVDD, \cite{perera2019learning} leverages a labeled dataset from a close domain to fine-tune pre-trained two CNNs in order to learn new \textit{good} features. The goodness of these features is quantified by the compactness (inter-class variance) for the target class and descriptiveness (cross-entropy) for the labeled dataset. Despite avoiding hyper-sphere collapse and outperforming OC-SVDD, this approach requires two pre-trained neural networks and a large labeled dataset along with the target dataset. Another approach that also makes use of a large auxiliary dataset is Outlier Exposure (OE) \cite{hendrycks2018deep}, which is a supervised approach that trains a standard neural network classifier while exposing it to a diverse set of non-examples on which the output of the classifier is optimized to follow a uniform distribution using another cross-entropy loss.
\section{Proposed Method}
\subsection{Architecture}
An undercomplete deep autoencoder is a type of unsupervised feed-forward neural network for learning a lower-dimensional feature representation of a particular dataset via reconstructing the input back at the output layer. To prevent autoencoders from converging to trivial solutions such as the identity mapping; a bottleneck layer with output $\mathbf{z}$ such that its dimension is less than the dimension of the input $\mathbf{x}$. The forward pass is computed as such:
\begin{equation}
    \mathbf{s} = \mathscr{E}(\mathbf{x}),
\end{equation}
\begin{equation}
    \mathbf{z} = \mathscr{Z}(\mathbf{s}),
\end{equation}
\begin{equation}
    \hat{\mathbf{x}}= \mathscr{D}(\mathbf{z}),
\end{equation}
where $\mathbf{x}$ is the input, $\mathscr{Z}$ is the bottleneck layer, $\mathscr{E}$ and $\mathscr{D}$ are convolutional neural networks representing the encoder and the decoder modules respectively. Typically, such models are trained to minimize the $\mathcal{L}_2$-norm of the difference between the input and the reconstructed output $\|\hat{\mathbf{x}}-\mathbf{x}\|_2$.
As previously discussed, the choice of the activation function of $\mathbf{z}$ plays an important rule in anomaly detection. Activation functions that quantize the latent code or encourage forming clusters are preferable. In our experiments, we find that confining the latent code to have values between $[-1,1]$ with a $\texttt{tanh}$ activatin function as we maximize the loss over the negative dataset during the latent-shaping phase has a regularizing effect. We also note that unbounded activation functions such as ReLU tend to have poor performance.

\subsection{Terminology}

\textbf{Positive Dataset ($\mathcal{D}^+$):} This is the dataset that contains the normal class(es), for example, the \textit{plane} class from CIFAR-10.\\
\textbf{Negative Dataset ($\mathcal{D}^-$):} This is a secondary unlabeled dataset containing negative examples from a similar domain as $\mathcal{D}^+$. The choice of $\mathcal{D}^-$ depends on $\mathcal{D}^+$. For example if $\mathcal{D}^+$ is the digit 0 from MNIST,  $\mathcal{D}^-$ might be random strokes or another dataset with similar features such as Omniglot\cite{tang2017vehicle}. It is important to note that the model should not be tested on  $\mathcal{D}^-$ since this violates the assumption of not knowing anomalies.\\
\textbf{Anomaly Dataset ($\mathcal{D}^a$):} This is a test dataset that contains classes that are neither in $\mathcal{D}^+$ nor in $\mathcal{D}^-$.

\noindent
\textbf{Feature Extraction Phase:} This is the first phase of training. The model is simply trained to reconstruct its input. \\
\textbf{Latent-Shaping Phase: }This is the second phase of training. The encoder and decoder networks are frozen and only the latent layer is active. 

\subsection{Training for Anomaly Detection}

Given a dataset $\mathcal{D}^+$ and a negative dataset $\mathcal{D}^-$ from a similar domain to $\mathcal{D}^+$, we divide the training process into two phases; the first phase is reconstructing samples from $\mathcal{D}^+$ by minimizing the loss function $\mathcal{L} = \|\mathbf{\hat{x}}^{+}-\mathbf{x}^+ \|_2$ until convergence, where $\mathbf{x}^+$ is the input drawn from $\mathcal{D}^+$ and $\mathbf{\hat{x}^+}$ is the output of the autoencoder. In the second phase, we freeze the model except for the latent layer and minimize the following loss function: 
\begin{equation}
    \mathcal{L} =\|\hat{{\mathbf{x}}}^+ -\mathbf{x}^+\|_2 + \beta \| \gamma - \|\hat{{\mathbf{x}}}^- -\mathbf{x}^-\|_2 \|_2
\end{equation}
where  $\mathbf{x^-}$ is a sampled batch from $\mathcal{D^-}$ , $\mathbf{\hat{x}^-}$ is its reconstruction, $\beta$ is a hyperparameter that controls the effect of the two parts of the loss function and $\gamma$ is another hyperparameter indicating that we are satisfied if the reconstruction error $\|\mathbf{\hat{x}^-} - \mathbf{x^-} \|_2$  of the negative dataset exceeds a certain value.

\begin{algorithm}
\DontPrintSemicolon
\SetAlgoLined
\KwIn{Positive ($\mathcal{D}^+$) and Negative ($\mathcal{D}^-$) datasets\\
\tcp*[l]{$\mathscr{E}$: Encoder, $\mathscr{Z}$: Latent Layer, $\mathscr{D}$: Decoder}}
\KwOut{Trained model}
\tcp*[l]{Feature extraction phase}
\tcp*[l]{Sample mini batches from $\mathcal{D}^+$}
\For{$\mathbf{x}^+ \in \mathcal{D}^+$ \textbf{until convergence}}
{
    $\hat{\mathbf{x}}^+ = \mathcal{D}(\mathscr{Z}(\mathscr{E}(\mathbf{x}^+))$\;
    $\mathcal{L} =  \|{\mathbf{\hat{x}^+}}-\mathbf{x^+}\|_2$\;
    \tcp*[l]{backpropagation step}
    Minimize $\mathcal{L}$\;
}

$FreezeEncoder()$\;
$FreezeDecoder()$\;
\tcp*[l]{Latent-shaping phase}
\tcp*[l]{Sample mini batches from $(\mathcal{D}^+,\mathcal{D}^-)$}
 \For{$(\mathbf{x}^+, \mathbf{x}^-) \in (\mathcal{D}^+,\mathcal{D}^-)$ \textbf{until convergence}}{
    $[\mathbf{z}^+, \mathbf{z}^-] = \mathscr{Z}(\mathscr{E}([\mathbf{x}^+,\mathbf{x}^-]))$\;
    $[\hat{\mathbf{x}}^+,\hat{\mathbf{x}}^-]=\mathcal{D}([\mathbf{z}^+, \mathbf{z}^-])$\;
    $\mathcal{L} =\|\hat{\mathbf{x}}^+-\mathbf{x}^+ \|_2 + \beta \| \gamma - \|\hat{\mathbf{x}}^- - \mathbf{x}^- \|_2 \|_2 $\;
    \tcp*[l]{backpropagation step}
    Minimize $\mathcal{L}$\;
}
\caption{Anomaly Detection Training}
\label{algo:alg1}
\end{algorithm}

\subsection{Predicting Anomalies}

We use reconstruction error $\mathcal{L}(x) =\|\mathbf{\hat{x}}-\mathbf{x} \|_2$ to distinguish between anomalies and normal data where $\mathbf{x}$ is the test sample and $\mathbf{\hat{x}}$ is the reconstructed output. More specifically, we set a threshold $\alpha$ such that if $\mathcal{L}(\mathbf{x}) > \alpha$ the output is considered to be anomalous.

\section{Theoretical
Justification}
\subsection{Formulation}
\newcommand*{\vertbar}{\rule[-1ex]{0.5pt}{2.5ex}}
\newcommand*{\horzbar}{\rule[.5ex]{2.5ex}{0.5pt}}
\newcommand{\centered}[1]{\begin{tabular}{l} #1 \end{tabular}}
In this section, we present theoretical justification for the reasoning behind selective freezing and the second phase loss function. We would like to show that the process described in algorithm 1 implies that the reconstruction loss for a latent-insensitive autoencoder $(\mathcal{L}_{LIS})$ remains equivalent to the reconstruction loss of a standard autoencoder $(\mathcal{L}_{AE})$ for normal (positive) samples but larger for anomalies. More formally, under certain assumptions for negative dataset ($\mathcal{D^-}$), $\mathcal{L}_{LIS}(\mathbf{x^+}) = \mathcal{L}_{AE}(\mathbf{x^+})$ and  $\mathcal{L}_{LIS}(\mathbf{x}^a) \ge \mathcal{L}_{AE}(\mathbf{x}^a)$ where $\mathbf{x}^a$ is an anomalous sample.\\
\noindent
From optimiality of autoencoders \cite{bourlard1988auto}, we know that absent any non-linear activation functions, a linear autoencoder corresponds to singular value decomposition (SVD); henceforth, we use SVD interchangeably with linear autoencoders. Given an $m \times n$ data matrix $\mathbf{X}^+$, 
we decompose $\mathbf{R}^m $ into $\mathbf{X^{||} \oplus X^{\perp}}$, where $\mathbf{X}^{||} := Col(X^+)$ and  $\mathbf{X}^{\perp}$ is its orthogonal complement $Null(\mathbf{X}^{+^T})$.

\vspace{.1cm}
\noindent
We further decompose $\mathbf{X^+}$ using SVD:
$$\mathbf{X^+ }= \mathbf{U} \mathbf{\Sigma} \mathbf{V}^T,$$
where $\mathbf{U}$ and $\mathbf{V}$ are orthonormal matrices and $\mathbf{\Sigma}$ is a diagonal matrix such that $\mathbf{\Sigma}= [\sigma_1  \hspace{3pt}... \hspace{3pt} \sigma_{rank}  \hspace{1pt}|  \hspace{1pt} 0 ]$. However, in practice it is rarely separated this neatly, specially when dealing with large number of samples of a high-dimensional dataset; therefore, we resort to reduced-SVD where we take the first $r$ columns of $\mathbf{U}$ with the caveat that the choice of $r$ is a hyper-parameter.

\[
\mathbf{U} = 
\left[
  \begin{array}{cccccc}
    \vertbar &      &\vertbar  & \vertbar&      &\vertbar \\
    \mathbf{u}_{1}    &\ldots& \mathbf{u}_{r}    &\mathbf{u}_{r+1}  &\ldots & \mathbf{u}_{n}   \\
    \vertbar &      &\vertbar  & \vertbar&       &\vertbar 
  \end{array}
\right]
\]
\\
The $\mathbf{U}$ matrix can be divided thus: $\mathbf{U} = [\mathbf{U}_r  |  \mathbf{U}_c]$, and from Eckart–Young low-rank approximation theorem, columns of $\mathbf{U}_r \approx Basis(\mathbf{X^{||}})$ and columns of $\mathbf{U}_c  \approx Basis(\mathbf{X^{\perp}})$.
\\
A linear autoencoder with $r$-dimensional latent layer is equivalent to the following transform:
$$\mathbf{\hat{x}} = \mathbf{U}_r \mathbf{U}_r^T \mathbf{x}$$
Where $\mathbf{U}_r$ and $\mathbf{U}_r^T$ represent the decoder and the encoder respectively. Furthermore, any data point $\mathbf{x} \in \mathbf{R}^m$ can be represented as $\mathbf{x} = \mathbf{U}_r\mathbf{z} + \mathbf{U}_c\mathbf{c}$, where $\mathbf{c}$ and $\mathbf{z}$ are ($m-r$) and $r$-dimensional real vectors. By orthonormality, we have the following identities:
$\mathbf{U}_r^T \mathbf{U}_r = \mathbf{I}_r$ and $\mathbf{U}_r^T \mathbf{U}_c = \mathbf{0}$, where $\mathbf{I}_r$ is an $r$-identity matrix. As a shorthand, we write $\|.\|$ instead of  $\|.\|_2^2$.  Using these two identities, we rewrite the reconstruction loss $\|\mathbf{\hat{x}}-\mathbf{x} \|$ as following:
\begin{align*}
    \mathbf{U}_r \mathbf{U}_r^T \mathbf{x} & = \mathbf{U}_r \mathbf{U}_r^T [\mathbf{U}_r \mathbf{z} +\mathbf{U}_c\mathbf{c}] =\mathbf{U}_r [\mathbf{U}_r^T \mathbf{U}_r\mathbf{z} +\mathbf{U}_r^T \mathbf{U}_c\mathbf{c}] =\mathbf{U}_r\mathbf{z}\\
    \mathcal{L}_{AE}(\mathbf{x}) & = \|\mathbf{U}_r \mathbf{U}_r^T \mathbf{x} - \mathbf{x} \| =\|\mathbf{U}_r \mathbf{z} -\mathbf{U}_r \mathbf{z} -\mathbf{U}_c \mathbf{c}\| = 
\| \mathbf{U}_c \mathbf{c} \| = \|\mathbf{c} \|
\end{align*}
We note that the loss function is agnostic to the nature of $\mathbf{c}$ and is only concerned with its magnitude. The assumption for anomaly detection under this setting is that $\|\mathbf{c}^a\| >\|\mathbf{c}^+\|$, where $\mathbf{c}^a$ and $\mathbf{c}^+$ correspond to orthogonal components for anomalies and positive data respectively. We posit that while this agnosticism is desirable for potential generality, it is not optimal for anomaly detection; hence, we modify the loss score to depend on the nature of $\mathbf{c}$:
\begin{equation*}
   \mathcal{L}_{LIS}(\mathbf{x}) = \|\mathbf{B}^T \mathbf{c}\| + \|\mathbf{c}\|
\end{equation*}
where $\mathbf{B}$ is an $r\times(m-r)$ matrix such that the loss is small for normal data but large for anomalies. In other words, we want $\|\mathbf{B}^T \mathbf{c}^{+}\| = 0$ and $\|\mathbf{B}^T \mathbf{c}^{a}\|$ to be large. \\
We define $\mathbf{C}^{||} :=$ orthonormal basis for $Col(\mathbf{C}^+)$ and $\mathbf{C}^{\perp} :=$ orthonormal basis for $Null(\mathbf{C}^{+^T})$, where $\mathbf{C^+}$ is the matrix of all positive orthogonal components $\mathbf{c^+}$.
We decompose $\mathbf{C}^{\perp}$ further into $\mathbf{C}^{-\perp}$ and $\mathbf{C}^{o\perp}$ where columns of $\mathbf{C}^{-\perp}$ are the basis of $Col(\mathbf{C}^{-})$ that are not in $\mathbf{C}^{||}$ and columns of $\mathbf{C}^{o\perp}$ are the remaining columns of $\mathbf{C}^{\perp}$.
\\
Since $\mathbf{R}^{m-r}= Col(\mathbf{C}^+) \oplus Col(\mathbf{C}^{-\perp}) \oplus Col(\mathbf{C}^{o\perp})$, any $\mathbf{c} \in \mathbf{R}^{m-r}$ can be written as $\mathbf{c} = \mathbf{C}^{||} \mathbf{p} + \mathbf{C}^{-\perp} \mathbf{q}+ \mathbf{C}^{o\perp} \mathbf{s}$, where $\mathbf{p}$, $\mathbf{q}$ and $\mathbf{s}$ are real vectors.\\
Despite the fact that we do not have access to $\mathbf{c}^{a}$, we can utilize other negative examples from similar domain and use $\mathbf{c^{-}}$ as a proxy for $\mathbf{c}^{a}$. Since $\mathbf{c}^{a}  = \mathbf{C}^{||} \mathbf{p} + \mathbf{C}^{-\perp} \mathbf{q}+ \mathbf{C}^{o\perp} \mathbf{s}$, maximizing $\|\mathbf{B}^T \mathbf{C}^{-\perp}\|$ implies maximizing $\|\mathbf{B}^T \mathbf{c}^{a}\|$ assuming that $\|\mathbf{C}^{-\perp} \mathbf{q}\| \neq 0$. The later assumption hinges on the fact that $\mathbf{X^-}$ is from a \textit{similar} domain. Therefore, we end up with the goal of finding $\mathbf{B}$ such that $\|\mathbf{B}^T \mathbf{c}^+\| = 0$ and $\|\mathbf{B}^T \mathbf{c}^-\|$ is large.\\
\begin{align*}
\hat{\mathbf{B}} &= \texttt{argmin} ( \|\mathbf{B}^T \mathbf{c}^+\| - \mathbf{\beta} \|\mathbf{B}^T \mathbf{c}^-\|)\\
&\centerline{where $\beta$ controls the importance of the second term.}\\
&= \texttt{argmin} ( \|\mathbf{U}_r \mathbf{B}^T \mathbf{c}^+\| - \mathbf{\beta} \|\mathbf{U}_r \mathbf{B}^T \mathbf{c}^-\|)\\
&\centerline{since orthonormal transformations preserve the dot product.} \\
&=\texttt{argmin} (\|\mathbf{U}_r \mathbf{B}^T \mathbf{c}^+\|  + \|-\mathbf{U}_c \mathbf{c}^+\| - \mathbf{\beta} \|\mathbf{U}_r \mathbf{B}^T \mathbf{c}^-\| -\beta \|-\mathbf{U}_c \mathbf{c}^-\| )\\
&\centerline{since adding constant to $\texttt{argmin}$ does not affect the objective.}\\
&=\texttt{argmin} (\|\mathbf{U}_r \mathbf{B}^T \mathbf{c}^+  - \mathbf{U}_c \mathbf{c}^+\| - \beta \|\mathbf{U}_r \mathbf{B}^T \mathbf{c}^- - \mathbf{U}_c \mathbf{c}^-\| )\\
&=\texttt{argmin} (\|\mathbf{U}_r \mathbf{z}^+ +\mathbf{U}_r \mathbf{B}^T \mathbf{c}^+ -\mathbf{U}_r \mathbf{z}^+ - \mathbf{U}_c \mathbf{c}^+\| \\
&\qquad\qquad-\mathbf{\beta} \|\mathbf{U}_r \mathbf{z}^+ + \mathbf{U}_r \mathbf{B}^T \mathbf{c}^- -\mathbf{U}_r \mathbf{z}^- - \mathbf{U}_c \mathbf{c}^-\| )\\
&=\texttt{argmin} (\|\mathbf{U}_r \mathbf{z}^+ +\mathbf{U}_r \mathbf{B}^T \mathbf{c}^+ -\mathbf{x}^+\|
- \mathbf{\beta} \|\mathbf{U}_r \mathbf{z}^- + \mathbf{U}_r \mathbf{B}^T \mathbf{c}^- -\mathbf{x}^-\| )\\
&=\texttt{argmin} (\|\mathbf{U}_r \mathbf{U}_r^T \mathbf{x}^+ +\mathbf{U}_r \mathbf{B}^T \mathbf{U}_c^T \mathbf{x}^+ -\mathbf{x}^+\| 
\\
&\qquad\qquad - \mathbf{\beta} \|\mathbf{U}_r \mathbf{U}_r^T \mathbf{x}^- +\mathbf{U}_r \mathbf{B}^T \mathbf{U}_c^T \mathbf{x}^- -\mathbf{x}^-\|)\\
&=\texttt{argmin} (\| \mathbf{U}_r (\mathbf{U}_r^T +\mathbf{B}^T\mathbf{U}_c^T)\mathbf{x}^+ -\mathbf{x}^+\| \\
&\qquad\qquad - \beta \| \mathbf{U}_r(\mathbf{U}_r^T + \mathbf{B}^T \mathbf{U}_c^T)\mathbf{x}^-  -\mathbf{x}^-\|)\\
\mathbf{E} &:= \mathbf{U}_r + \mathbf{U}_c \mathbf{B}\\
\mathbf{\hat{E}} &=\texttt{argmin} (\| \mathbf{U}_r \mathbf{E}^T \mathbf{x}^+  -\mathbf{x^+}\| - \mathbf{\beta} \| \mathbf{U}_r\mathbf{E}^T \mathbf{x}^-  -\mathbf{x}^-\|)
\end{align*}
\noindent
In practice, we cannot maximize $\|\mathbf{U}_r\mathbf{E}^T\mathbf{x}^-  -\mathbf{x}^-\|$ indefinitely and we are satisfied if it reaches a 
certain large $\gamma$:
\begin{equation*}
    \mathbf{\hat{E}} =\texttt{argmin} (\| \mathbf{U}_r \mathbf{E}^T \mathbf{x}^+  -\mathbf{x}^+\| + \beta \| \gamma - \|\mathbf{U}_r \mathbf{E}^T \mathbf{x}^-  -\mathbf{x}^-\| \|)
\end{equation*}

\begin{figure}
\raggedright

\begin{subfigure}[b]{0.75\linewidth}
\begin{tabular}
      {ll}
      \begin{turn}{90}
      \makebox[1.5cm]{
      \begin{turn}{-90}
      (a) 
      \end{turn}}\end{turn}
      &
      \includegraphics[trim={0cm 8.5cm 0cm 0cm},clip, width=\linewidth]{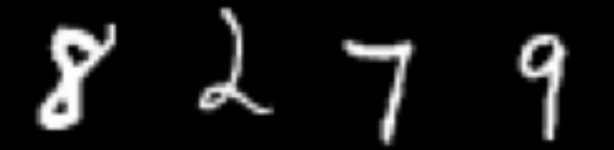}
 \end{tabular}
  \label{fig:mnist1} 
\end{subfigure}

\begin{subfigure}[b]{0.75\linewidth}
\begin{tabular}
      {ll}
      \begin{turn}{90}
      \makebox[1.5cm]{
      \begin{turn}{-90}
      (b) 
      \end{turn}}\end{turn}
      &
   \includegraphics[trim={0cm 6.8cm 0cm 1.7cm},clip, width=\linewidth]{MNIST_Orth_Samp.pdf}
      \\
 \end{tabular}
  \label{fig:mnist2} 
\end{subfigure}

\begin{subfigure}[b]{0.75\linewidth}
\begin{tabular}
      {ll}
      \begin{turn}{90}
      \makebox[1.5cm]{
      \begin{turn}{-90}
      (c) 
      \end{turn}}\end{turn}
      &
   \includegraphics[trim={0cm 5.1cm 0cm 3.4cm},clip, width=\linewidth]{MNIST_Orth_Samp.pdf}
      \\
 \end{tabular}
  \label{fig:mnist3} 
\end{subfigure}

\begin{subfigure}[b]{0.75\linewidth}
\begin{tabular}
      {ll}
      \begin{turn}{90}
      \makebox[1.5cm]{
      \begin{turn}{-90}
      (d) 
      \end{turn}}\end{turn}
      &
   \includegraphics[trim={0cm 3.4cm 0cm 5.1cm},clip, width=\linewidth]{MNIST_Orth_Samp.pdf}
      \\
 \end{tabular}
  \label{fig:mnist4} 
\end{subfigure}

\begin{subfigure}[b]{0.75\linewidth}
\begin{tabular}
      {ll}
      \begin{turn}{90}
      \makebox[1.5cm]{
      \begin{turn}{-90}
      (e) 
      \end{turn}}\end{turn}
      &
   \includegraphics[trim={0cm 1.7cm 0cm 6.8cm},clip, width=\linewidth]{MNIST_Orth_Samp.pdf}
      \\
 \end{tabular}
  \label{fig:mnist5} 
\end{subfigure}

\begin{subfigure}[b]{0.75\linewidth}
\begin{tabular}
      {ll}
      \begin{turn}{90}
      \makebox[1.5cm]{
      \begin{turn}{-90}
      (f) 
      \end{turn}}\end{turn}
      &
   \includegraphics[trim={0cm 0cm 0cm 8.5cm},clip, width=\linewidth]{MNIST_Orth_Samp.pdf}
      
 \end{tabular}
  \label{fig:mnist6} 
\end{subfigure}


\caption{Comparison between Linear AE and LIS-AE. Digit-8 is the normal task. (a) Inputs. (b) Orthogonal vector. (c) AE reconstruction of orthogonal vectors(zero images). (d) LIS-AE reconstruction  of orthogonal vectors. (e) AE reconstruction of inputs. (f) LIS-AE reconstruction of inputs.}
\end{figure}

We notice that in order for this to work, the decoder $\mathbf{U}_r$ has to be known and remain fixed (frozen). This suggests a two-phase training where we first compute the decoder and encoder networks, and in the second phase the decoder is fixed while the encoder $\mathbf{E}^T$ is modified using the new loss. In fig. 4, a linear version of LIS-AE is trained on digit-8 from MNIST with Omniglot as a negative dataset. We perform orthogonal decomposition on each input by projecting it onto digit-8 subspace to get its projection and orthogonal vectors.
We then feed each vector separately to a regular linear AE and linear LIS-AE. We observe that the regular autoencoder outputs zero images for the orthogonal part of each sample regardless of the class it belongs to. However, in the case of LIS-AE, it behaves differently for normal class than for anomalous classes.

      


We also notice that orthogonal projections do not form a semantically meaningful representation in pixel space. In order to gain a better representation we use a deep AE. For this non-linear case, we treat the middle part of the network as an \textit{inner} linear autoencoder which is operating on a more semantically meaningful transformed version of the data. 
This suggests a stacked autoencoder archiecture where another loss term for the \textit{inner} autoencoder is added in the first phase to make sure that the output of the layer after latent is similar to the latent input. In the second phase we freeze the entire network except for the encoder of the \textit{inner} autoencoder (latent layer of entire model) and minimize the reconstruction error of positive examples while maximizing the loss for negative examples.
However, in our experiments we observed that adding these loss terms was not necessary and a similar loss to the linear case produced similar results since we are only considering reconstruction scores of the outer model. Therefore, we keep the entire network frozen except for the latent layer while directly minimizing the following loss as before. (eq. 4)
\subsection{Intuition}
For concreteness, we consider the following simple, supervised case. Given a dataset $\mathbf{X^+}$ such that for each $\mathbf{x^+} \in \mathbf{X^+}$:
\begin{equation*}
\mathbf{x^+} \sim \mathcal{N}\left( \mathbf{\mu}=0,\mathbf{\Sigma}=
\begin{bmatrix}
1 & 0 & 0\\
0 & 0.2 & 0\\
0 & 0 & 0.01
\end{bmatrix}
\right)
\end{equation*}
\noindent
We notice that most of the variance in data is along the x-axis. Training a linear autoencoder with latent dimension $r=1$, results in $\mathbf{D}^T = \begin{bmatrix} 1 & 0 & 0 \end{bmatrix}$ and $\mathbf{E}^T = \begin{bmatrix} 1 & 0 & 0 \end{bmatrix}$ where $\mathbf{D}$ and $\mathbf{E}$ are the decoder and encoder networks respectively.\\
Given input $\mathbf{x}^T = \begin{bmatrix} x & y & z \end{bmatrix}$,
\begin{equation*}
\mathbf{\tilde{x}} = \mathbf{D} \mathbf{E}^T\mathbf{x} = 
\begin{bmatrix} 1 \\ 0 \\ 0 \end{bmatrix}\begin{bmatrix} 1 & 0 & 0 \end{bmatrix}
\begin{bmatrix} x\\ y \\ z \end{bmatrix} = \begin{bmatrix} x\\ 0 \\ 0 \end{bmatrix},
\end{equation*}
the loss score is $\mathbf{L} = \|\mathbf{x} -\mathbf{\tilde{x}}\|^2 = y^2 + z^2$. Training a LIS-AE on negative samples that have only nonzero values along the z-axis, we end up with the same $\mathbf{D}$ and a modified $\hat{\mathbf{E}}^T = \begin{bmatrix} 1 & 0 & \gamma \end{bmatrix}$, where $\gamma$ is a large number and then
\begin{equation*}
\mathbf{\hat{\tilde{x}}} = \mathbf{D} \hat{\mathbf{E}}^T\mathbf{x} = 
\begin{bmatrix} 1 \\ 0 \\ 0 \end{bmatrix}\begin{bmatrix} 1 & 0 & \gamma \end{bmatrix}
\begin{bmatrix} x\\ y \\ z \end{bmatrix} = \begin{bmatrix} x+\gamma z\\ 0 \\ 0 \end{bmatrix},
\end{equation*}
that results in 
$\hat{\mathbf{L}} = \|\mathbf{x} -\mathbf{\hat{\tilde{x}}} \| \approx y^2 + \gamma^2 z^2$ with 
$\mathbf{x^+}$ has the form $\begin{bmatrix} x_+ & y_+ & 0 \end{bmatrix}^T$,
$\mathbf{x^a}$ has the form $\begin{bmatrix} x_a & y_a & z_a \end{bmatrix}^T$ where $y_a, z_a \neq 0$. The new loss scores for $\mathbf{x^+}$ and $\mathbf{x^a}$ are: $$\mathbf{L}(\mathbf{x^+}) = y_+^2,\hspace{.1cm} \mathbf{L}(\mathbf{x^a}) = y_a^2 + \gamma^2 z_a^2$$
\noindent
In the case of regular Linear-AE (PCA), given $\mathbf{x}^+=(x_+,y_+, 0)$, for each point $(x_a,y_a,z_a)\in$ the cylinder: $\left(\sqrt{ y_a^2 + z_a^2} \leq y_+, x_a\in \mathbf{R}\right)$ the following holds: $\mathbf{L}(\mathbf{x^a}) = \mathbf{L}(\mathbf{x^+})$, making the two samples \textit{indistinguishable}. 
In the case of LIS-AE, $\hat{\mathbf{L}}(\mathbf{x^a}) = \hat{\mathbf{L}}(\mathbf{x^+})$ holds only for the elliptic cylinder $\left(\sqrt{ y_a^2 + \frac{z_a^2}{1/\gamma^2}} \leq y_+, x_a\in \mathbf{R}\right)$, and since $\gamma$ is a large number, the cross-section of the cylinder is \textit{squashed} in the z dimension resulting in heavily penalized loss in the z dimension but a regular loss in the y dimension. In this case, the two samples become \textit{indistinguishable} only for very small values of $z_a$.\\
We note that the new $\hat{\mathbf{E}}$ is merely a rotated and stretched version of the old $\mathbf{E}$ in the $xz$-plane. Thus, we can think of Linear LIS-AE as a regular PCA with its eigenvectors (columns of $\mathbf{U_r}$) \textit{stretched} and \textit{tilted} in the directions of the orthogonal complement of the eigenspace. This is done in such a way that keeps the column space of normal examples invariant under the new transformation $ \mathbf{U}_r \mathbf{E^T}$. By itself, this formulation remains ill-posed since there are infinite number of solutions that do not necessarily help with anomaly detection. More formally, given $\mathbf{E}:= \mathbf{U}_r + \mathbf{U}_c \mathbf{B}$, we can choose any matrix $\mathbf{B}$ such that $Null(\mathbf{B}^T) = Col(\mathbf{U_c^T X^+})$ since $ \mathbf{U}_r \mathbf{E^T}\mathbf{x}^+ =  \mathbf{U}_r (\mathbf{U}_r^T \mathbf{x}^+ + \mathbf{B}^T \mathbf{U}_c^T \mathbf{x}^+) = \mathbf{U}_r \mathbf{U}_r^T \mathbf{x}^+ \approx \mathbf{x}^+$. However, this does not guarantee any advantage for anomaly detection on similar data, even worse in practice, this modification process might result in a slightly worse performance if done arbitrarily since  the model usually has to sacrifice some extreme samples from the normal data to balance the two losses. Thus, the negative dataset is used to properly determine the directions of the \textit{tilt} and hyperparamteres ($\gamma$ and $\beta$) determine the importance and amount of stretching (or shrinking) without changing the normal case as much as possible. For deep LIS-AE, the same analogy holds albeit in a \textit{latent space}.

Deep architectures are not only useful for learning good representation, but can learn a non-linear transformation with useful properties for our objective such as linear separability of negative and positive samples. By adding a standard binary cross-entropy loss before the non-linear activation of the latent layer during the first phase, we ensure that the input of the latent layer is linearly-separable for positive and negative examples during the second phase. This linearly-separable variant (LinSep-LIS-AE) almost always performs better than directly using LIS-AE. We investigate the effect of this property on the second phase in section 5.2.
\section{Experiments}

\noindent
We report results on the following datasets, MNIST \cite{lecun-mnisthandwrittendigit-2010}, Fashion-MNIST \cite{xiao2017fashion}, SVHN \cite{netzer2011reading} and CIFAR-10 \cite{krizhevsky2009learning}. Results of our approach are compared to baseline models with the same capacity for autoencoder-based methods.

\subsection{Anomaly Detection}
\begin{table}
\caption{Average AUC for 10 tasks sampled from MNIST, Fashion-MNIST and 5 2-class tasks sampled from MNIST.}
\label{T1}
    \begin{tabular}{c|ccc}
    \toprule
    Model & MNIST & Fashion-MNIST&2-Class MNIST \\
 \midrule
    KDE                    &0.9568 &0.9183& 0.9206\\
    IF                     &0.8624 &0.9144& 0.73018\\
    OC-SVM                 &0.9108 &0.8608&0.8741\\
    OC-DSVDD               &0.9489 &0.8577&0.8972\\ 
    \midrule        
    AnoGAN               &0.9579 &0.9098&0.8406\\
    AnoGAN-FM          &0.9544 &0.9072&0.8353\\
    Linear-AE              &0.9412  &0.8845 &0.8915\\
    VAE                    &0.9642  &0.9092 &0.9263\\
    Mem-AE                 &0.9714  &0.9131 &0.9352\\
    Sig-RepNN (N=4)        &0.9661  &0.9124 &0.9261\\
    AE                     &0.9601  &0.9076 &0.9221\\
    LIS-AE                 &\bf0.9768  &\bf0.9256&\bf0.9457\\
    \bottomrule
    \end{tabular}
\end{table}
In this section, we test LIS-AE for anomaly detection on image data in unsupervised settings. Given a standard classification dataset, we group a set of classes together into a new dataset and consider it the \textit{"normal"} dataset. \textit{The rest of classes that are not in the normal nor in the negative datasets are considered anomalies}. During training, our model is presented only with the normal dataset and the additional negative dataset. We evaluate the performance on test data comprised of both the "normal" and "anomalous" groups.\\ For MNIST and Fashion-MNIST, the encoder network consists of two Convolutional layers with LeakyReLU non-linearities followed by a fully-connected bottleneck layer with \textit{tanh} activation function. The decoder network consists of a fully-connected layer followed by a LeakyReLU and two Deconvolution layers with LeakyReLU activation functions and a final convolution layer with sigmoid situated at the final output. For SVHN and CIFAR-10 we use latent layer with larger sizes and higher capacity networks with same depth. It is worth noting that the choice of latent layer size has the most effect on performance for all models (compared to other hyper-parameters). We report the best performing latent dimension for all models.
\\
In table (1), we compare LIS-AE with several autoencoder-based anomaly detection models as baselines, all of which share the exact same architecture. It is worth noting that the most direct comparison is between LIS-AE and AE since not only they have the same architecture, they have the exact same encoder and decoder weights and their performance is merely measured before and after the latent-shaping phase. We use a different variant of RepNN with a sigmoid activation function $\sigma(x) = 1/(1+\texttt{exp}(-x))$ placed before the $\texttt{tanh}$ staircase function approximation described in section 2. This is mainly used because \textit{"squashing"} the input between 0 and 1 before passing it to the staircase function gives more robust and easy-to-train network. We only report the best results for Sig-RepNN with 4 activation levels. For anomaly GAN (AnoGAN) \cite{schlegl2017unsupervised}, we follow the implementation described in \cite{schlegl2019f}. We train a W-GAN \cite{gulrajani2017improved} with gradient penalty and report performance for two anomaly scores, namely, encoder-generator reconstruction loss and additional feature-matching distance score in the discriminator feature space (AnoGAN-FM). 

\begin{table}
\label{T2}
\caption{Average AUC for 10 anomaly detection tasks sampled from SVHN and CIFAR-10 are shown.}
\begin{tabular}{c|cc}
  \toprule
  Model & SVHN & CIFAR-10 \\
    \midrule
    KDE                    &0.5648 &0.5752\\
    IF                     &0.5112 &0.6097\\
    OC-SVM                 &0.5047 &0.5651\\
    OC-DSVDD               &0.5681 &0.6411 \\ 
    \midrule
    AnoGAN                 &0.5598&0.5843\\
    AnoGAN-FM              &0.5645&0.5880\\
    Linear-AE              &0.5702 &0.5753\\
    VAE                    &0.5692 &0.5734\\
    Mem-AE                 &0.5720 &0.5931\\
    Sig-RepNN (N=4)        &0.5684 &0.5719\\
    AE                     &0.5698 &0.5703\\
    LIS-AE       &\bf0.6886  &\bf0.8145\\
    LinSep-LIS-AE &\bf0.7701 &\bf0.8858\\
    \midrule
    Sup. LIS-AE &0.7573 &0.8384\\
    Sup. LinSep-LIS-AE &0.8479 &0.9170\\
\bottomrule 
\end{tabular}
\end{table}

For AnoGAN, Linear-AE, AE, VAE, RepNN, MemAE, and LIS-A, we use reconstruction error $\mathcal{L}(\mathbf{x})$
such that if $\mathcal{L}(\mathbf{x}) > \alpha$ the input is considered an anomaly. 
Varying the threshold $\alpha$, we are able to compute the area under the curve (AUC) as a measure of performance. Similarly, for OC-SVDD (equivalently OC-SVM with rbf kernel) and OC-DSVDD, we vary inverse length scale $\gamma$ and use predicted class label. For Kernel density estimation (KDE) \cite{parzen1962estimation}, we vary the threshold $\alpha$ over the log-likelihood scores. For Isolation Forest (IF) 
\cite{liu2008isolation}, we vary the threshold $\alpha$ over the anomaly score calculated by the Isolation Forest algorithm.  

\begin{figure}
    \centering
    \includegraphics[trim={2.5cm 2.5cm 2.5cm 2.5cm}, width=.5\textwidth]{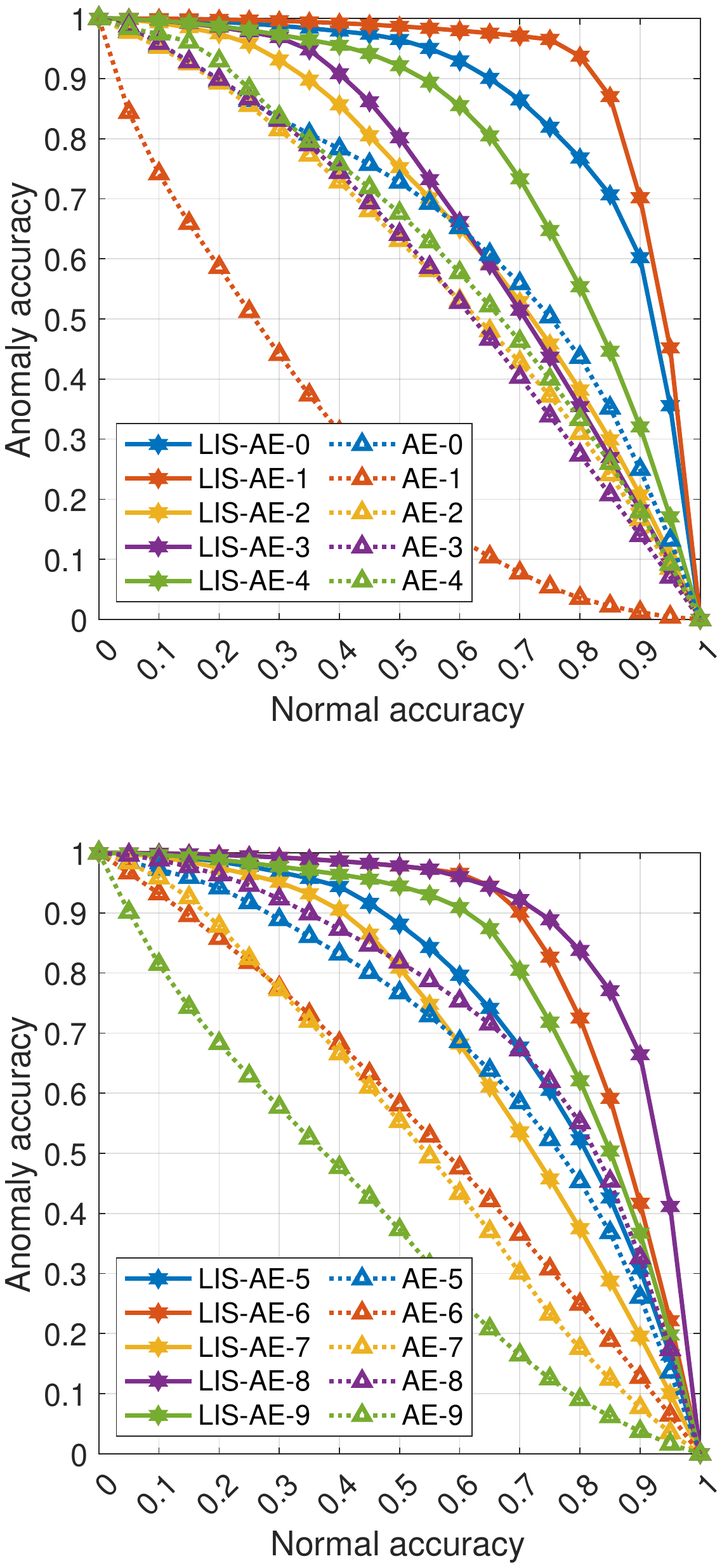}
    \caption{Each line represents a trade-off between accuracy of anomalies and normal data for CIFAR10. The top pane shows accuracies on tasks 0-4 and the bottom shows accuracies on tasks 5-9. Note that as the threshold value $\alpha$ increases the model favors accepting anomalies over misclassifying normal examples. LIS-AE gives a significant margin compared to base AE.}
    \label{fig:CIFAR10_accuracies}
\end{figure}

\begin{figure}
    \begin{tabular}{ll}
     \includegraphics[trim={.05cm .1cm 24.7cm .2cm},clip, width=.475\linewidth]{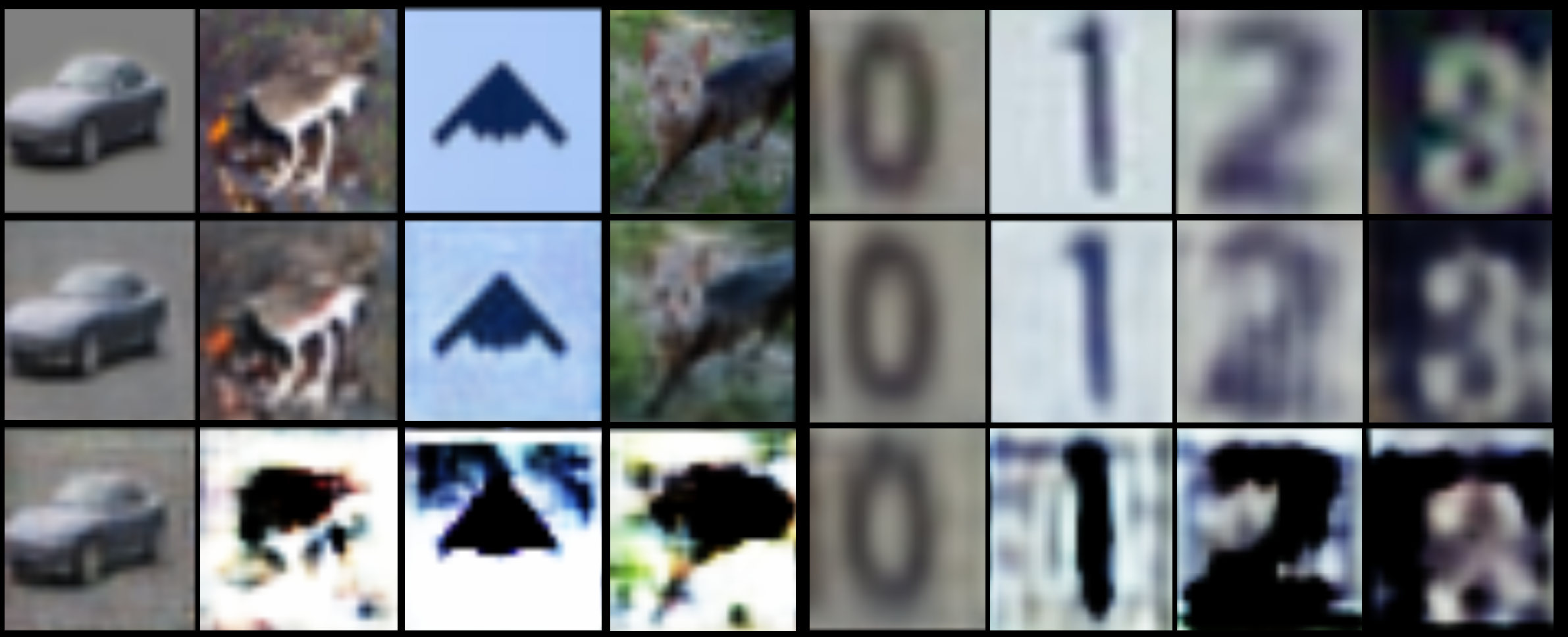}&
     \includegraphics[trim={25.9cm .1cm .35cm .2cm},clip, width=.45\linewidth]{combined.png}
    \end{tabular}
    
    \caption{Top row is test input from CIFAR-10 and SVHN, middle row is the output of a standard AE (first phase) and the bottom row is the output of LIS-AE. Trained on normal \textit{"car"} and \textit{"digit-0"} classes, LIS-AE only reconstructs samples of \textit{normal} class correctly.}
    
    \label{fig:cifar10_fig}
\end{figure}

The datasets tested in table (1) are MNIST and Fashion-MNIST. To train LIS-AE on MNIST we use Omniglot \cite{lake2015human} as our negative dataset since it shares similar compositional characteristics with MNIST. Since Omniglot is a relatively small dataset, we diversify the negative examples with various augmentation techniques, namely, Gaussian blurring, random cropping, horizontal and vertical flipping. We test two settings for MNIST, a 1-class setting where normal dataset is one particular class and the rest of the dataset classes are considered anomalies. The process is repeated for all classes and averge AUC for 10 classes is reported. Another setting is 2-class MNIST where the normal dataset consists of two classes and the remaining classes are considered anomalies. For example, the first task contains digits 0 and 1 and the remaining digits are considered anomalies, the second task contains digit 2 and 3, and so forth. This setting is more challenging since there is more than one class present in the normal dataset. 
For Fashion-MNIST, the choice of negative example is different. We use the \textit{next class} as the negative dataset and we do not include it with anomalies (i.e. remaining classes) during test time.
\\
We note that LIS-AE achieves superior performance to all compared approaches, however, we also notice that these settings are comparatively easy and all tested models performed adequately including classical non-deep approaches.
\noindent
In table (2), we show performance on SVHN and CIFAR-10 which are more complex dataests compared to MNIST and Fashion-MNIST. To train LIS-AE, we split each dataset into two datasets, each split is used as negative examples for the other one. Note that we only test on the remaining classes which are not in the normal nor the negative datasets. For example, the first dataset from CIFAR-10 has five classes, namely, airplane, automobile, bird, cat and deer while the second one has dog, frog, horse, ship and truck.
Training on airplane as the first normal task, LIS-AE maximizes the loss for samples drawn from the negative dataset (dog, frog, ship and so forth). We then test its performance on airplane as the normal class and only on automobile, bird, cat and deer as anomalies. Note that we do not test on dog, frog and other classes in the negative dataset. This processes is repeated for all 10 classes and average AUC is reported. As mentioned in section 4.2, We introduce LinSep-LIS-AE as an improvement over base standard LIS-AE. The difference between the two models is only in the first phase where a binary cross-entropy loss is added to ensure that positive and negative examples are linearly seperabable during the second phase. The last two entries of the table are supervised upper bounds for each variant where the negative dataset is the same as outliers. In figure (\ref{fig:cifar10_fig}), we see that standard AE is prone to generalize so well for other classes which is not a desired property for anomaly detection. In contrast, LIS-AE only reconstructs \textit{normal} data faithfully which translates to the large performance gap we see in figure (\ref{fig:CIFAR10_accuracies}).

We also notice that despite CIFAR-10 being more complex than SVHN, most reconstruction-based models are performing better on CIFAR-10 than on SVHN. This is due to the fact that the difference between SVHN classes in terms of reconstruction is not as \textit{large} since they share similar compositional features and do appear in samples from other classes while for CIFAR-10 classes vary significantly. (e.g. digit-2 and digit-3 on a wall vs truck and bird)  

\subsection{Ablation}

\begin{table}[]
\caption{Average AUC for 10 anomaly detection tasks sampled from two 5-class MNIST, Fashion-MNIST, SVHN and CIFAR-10 datasets where a regular LIS-AE is trained with different negative Splits.
}

\begin{tabular}{@{}c|cccc@{}}
\midrule
Negative & \multicolumn{4}{c}{Positive   Data} \\
\hspace{.25em}Data & MNIST & Fashion & SVHN & CIFAR-10 \\
\midrule
None     & 0.9485  & 0.8740   & 0.5698 & 0.5703        \\
Omni     & 0.9605  & 0.9013   &   -   &   -       \\
MNIST    & 0.9778  & 0.8942   &   -   &   -        \\
Fashion  & 0.9482  & 0.9106   &   -   &   -         \\
\midrule
SVHN &    -   &     -    &0.6886            &0.7065         \\
CIFAR-10& -  &   - & 0.5481              & 0.8145        \\
\midrule
Same (Sup.)&0.9901    &0.9623    & 0.7573 & 0.8384  \\

\midrule
\end{tabular}
\end{table}

In this section, we investigate the effect of the nature of negative dataset and linear separability of positive and negative examples. In table (3) we train LIS-AE on different negative and positive datasets. Similar to table (2), we split each positive dataset into two datasets and follow the same settings as before with the exception of "\textit{None}" and "\textit{Supervised} " cases. The "\textit{None}" case indicates that no negative examples have been used whereas the \textit{Supervised} indicates that both outliers and negative datasets share the same classes. Note that this case is different from the case where the positive and negative datasets come from the same dataset.
Unless stated otherwise, we only test on classes (outliers) that are not in the positive nor in the negative datasets. For example, when MNIST is used as a source for both positive and negative datasets, the positive data starts with class 0 and negative dataset consists of classes 5 to 9 where the outliers are classes 1 to 4. This process is repeated for all 10 classes present in each dataset and average AUC is reported. Overall, using a negative dataset resulted in a significant increase in performance in every case  except for two important cases, namely, when Fashion-MNIST and CIFAR-10 were used as negative datasets for MNIST and SVHN respectively. This could be explained by the fact that the model was not capable of reconstructing Fashion-MNIST and CIFAR-10 classes in the first place. Moreover, shaping the latent layer in such a way that maximizes the loss for Fashion-MNIST and CIFAR-10 classes does not guarantee any advantage for anomaly detection of similar digit classes present in MNIST and SVHN. This coupled with the fact that this process in practice forces the model to ignore some samples from the normal dataset to balance the two losses which results in the performance degradation we observe in these two cases.



Table (4) is an excerpt of the complete table in the appendices where we examine the effect of each class present in the negative dataset on anomaly detection performance for other test classes from the CIFAR-10 dataset. We split CIFAR-10 into two separate datasets, the first split is used for selecting classes as negative datasets and the other split is used as outliers. For each class in CIFAR-10 we train eight models in different settings, the first setting is \textit{None} where we train a standard autoencoder with no negative examples as the base model. The remaining seven settings differ in the second phase, we select one class as our negative dataset and test the model performance on each individual class from the outlier dataset. The \textit{combined} setting is similar to the setting described in section 5.1 where we combine all negative classes in one 5-class negative dataset. Note that these classes are not the same as the classes in the outlier test dataset except for the final setting, which is an upper-bound supervised setting where the negative dataset is comprised of classes that are in the outlier dataset except for the positive class. This process is then repeated for all 10 classes in CIFAR-10.
Overall, we observe a significant performance increase over the base model with the general trend of negative classes significantly increasing anomaly detection performance for similar outliers. For example, the \textit{dog class} drastically improves performance on the cat class but not so much for the \textit{plane class}. However, we also notice two important exceptions, namely, when the \textit{horse class} is used as the negative dataset for the \textit{car class}, we notice a significant performance increase for the relatively similar \textit{deer class} as expected, however, when the \textit{horse class} is used as the negative dataset for the same \textit{deer class}, we notice that the performance does not improve as in the first case and even degrades for the care class.
\begin{table}[]
\caption{AUC for LIS-AE trained on individual positive and negative classes is reported. }
\begin{tabular}{@{}c|c|ccccc|c@{}}
\midrule
Positive & Negative&\multicolumn{5}{c}{Outliers}& \\
\hspace{.25em}Class &\hspace{.25em}Class & Plane & Car  & Bird & Cat  & Deer & avg\\
\midrule
\multirow{8}{*}{Car}    & None                      & 0.32  & -    & 0.34 & 0.33 & 0.33 & 0.330                \\
                          & Dog 5                     & 0.67  & -    & 0.89 &\bf0.93 & 0.90 & 0.848                \\
                          & Frog 6                    & 0.58  & -    &\bf0.90 & 0.90 & 0.91 & 0.823                \\
                          & Horse 7                   & 0.66  & -    & 0.88 & 0.90 &\bf0.92 & 0.840                \\
                          & Ship 8                    &\bf0.83& -    & 0.59 & 0.51 & 0.50 & 0.608                \\
                          & Truck 9                   & 0.51  & -    & 0.44 & 0.49 & 0.44 & 0.470                \\
                          & Comb. (5-9)            & 0.81  & -    & 0.92 & 0.92 & 0.94 & 0.898                \\
                          & Sup. (0-4)          & 0.89  & -    & 0.93 & 0.90 & 0.95 & 0.918                \\
\midrule
\multirow{8}{*}{Deer}   & None                      & 0.56  & 0.80 & 0.52 & 0.54 & -    & 0.605                \\
                          & Dog  5                  & 0.72  & 0.85 & \bf0.63 & \bf0.80 & -    & 0.750                \\
                          & Frog  6                 & 0.66  & 0.86 & 0.60 & 0.75 & -    & 0.718                \\
                          & Horse 7                 & 0.71  & \textcolor{red}{0.58}& 0.58 & 0.71 & -    & 0.645                \\
                          & Ship   8                  &\bf0.93  & 0.94 & \bf0.63 & 0.72 & -    & 0.805                \\
                          & Truck 9                   & 0.84  &\bf0.97 & 0.62 & 0.72 & -    & 0.773                \\
                          & Comb. (5-9)            & 0.87  & 0.95 & 0.61 & 0.73 & -    & 0.790                \\
                          & Sup. (0-4)          & 0.93  & 0.97 & 0.63 & 0.72 & -    & 0.813            \\  
\midrule
\end{tabular}
\end{table}
Other notable examples of this observation can be found in the appendices where, for instance, the \textit{dog class} improves performance on cat outliers, but causes noticeable degradation when used as the negative dataset for the same cat class. The gained performance, in the first case, is due to the fact that these classes share similar compositional features and backgrounds. However in the second case, the same property makes it difficult to balance the minimization and maximization loss during the latent-shaping phase. For example, car and truck images are very similar in this scenario that minimizing and maximizing the loss at the same time becomes contradictory. As posited in section 4.2, we mitigate this issue by adding a binary cross-entropy loss while training in the first phase to ensure that the input of the latent layer is linearly-separable for positive and negative examples. Notice that unlike other approaches \cite{hendrycks2018deep, perera2019learning}, this does not require a labeled positive or negative dataset and relies only on the fact that we have two distinct datasets. This linear separablity makes the second phase of training relatively easier and less contradictory. In table (5), we see that LinSep-LIS-AE mitigates this issue for the aforementioned cases and gives the AUC increase we observed in table (2).




\noindent


\begin{table}[]
\caption{AUC for LinSep-LIS-AE trained on individual positive and negative classes is reported.
}

\begin{tabular}{@{}c|c|ccccc|c@{}}
\midrule
Positive & Negative&\multicolumn{6}{c}{Outliers} \\
\hspace{.25em}Class &\hspace{.25em}Class & Plane & Car  & Bird & Cat  & Deer & avg\\
\midrule
\multirow{8}{*}{Car}    & None                      & 0.32  & -    & 0.34 & 0.33 & 0.33 & 0.330                \\
                          & Dog 5                     & 0.67  & -    & 0.94 & \bf0.97 & 0.95 & 0.883                \\
                          & Frog 6                    & 0.58  & -    & 0.93 & 0.96 & 0.96 & 0.858                \\
                          & Horse 7                   & 0.69  & -    & \bf0.95 & \bf0.97 & \bf0.97 & 0.895                \\
                          & Ship 8                    & \bf0.90  & -    & 0.78 & 0.79 & 0.76 & 0.808                \\
                          & Truck 9                   & 0.59  & -    & 0.77 & 0.82 & 0.73 & 0.728                \\
                          & Comb. (5-9)            & 0.90  & -    & 0.97 & 0.97 & 0.98 & 0.955                \\
                          & Sup. (0-4)          & 0.95  & -    & 0.98 & 0.98 & 0.98 & 0.9725               \\
                           \midrule
\multirow{8}{*}{Deer}   & None                      & 0.56  & 0.80 & 0.52 & 0.54 & -    & 0.605                \\
                          & Dog 5                     & 0.67  & 0.86 & \bf0.71 & \bf0.89 & -    & 0.783                \\
                          & Frog 6                    & 0.68  & 0.87 & 0.62 & 0.79 & -    & 0.740                \\
                          & Horse 7                   & 0.70  & 0.84 & 0.61 & 0.72 & -    & 0.718                \\
                          & Ship 8                    & \bf0.94  & 0.95 & 0.63 & 0.73 & -    & 0.813                \\
                          & Truck 9                   & 0.84  & \bf0.97 & 0.61 & 0.76 & -    & 0.795                \\
                          & Comb. (5-9)            & 0.90  & 0.97 & 0.66 & 0.80 & -    & 0.833                \\
                          & Sup. (0-4)          & 0.97  & 0.98 & 0.76 & 0.83 & -    & 0.885             \\
\midrule
\end{tabular}
\end{table}

\section{Conclusion}

In this paper we introduced a novel autoencoder-based model called Latent-Insensitive autoencoder (LIS-AE). With the help of negative samples drawn from a similar domain as the normal data we tune the weights of the bottleneck part of a standrad autoencoder such that the resulting model is able to reconstruct the target task while penalizing anomalous samples. We also presented theoretical justification for the reasoning behind our two-phase training process and the latent-shaping loss function along with a more powerful variant. Multiple ablation studies were conducted to explain the effect of negative classes and highlight other important aspects of our model. We tested our model in a variety of anomaly detection settings with multiple datasets of varying degrees of complexity. Experimental results showed significant performance improvement over compared methods. Future research will focus on possible ways for synthesizing negative examples for domains with limited data. We also hope to further study and employ various manifold learning approaches for latent space representation. 

\section*{Acknowledgement}
Artem Lenskiy was funded by Our Health in Our Hands (OHIOH), a strategic initiative of the Australian National University, which aims to transform healthcare by developing new personalised health technologies and solutions in collaboration with patients, clinicians, and health care providers.




\bibliographystyle{ACM-Reference-Format}
\bibliography{main}
\onecolumn
\appendix

\section{Effect of individual classes as negative examples}
As discussed in section 5.2, we examine the effect of each class present in the negative dataset on anomaly detection performance for other test classes from the CIFAR-10 dataset. The first table shows results for standard LIS-AE while the second table shows results for LinSep-LIS-AE.
\\\\
\begin{tabular}{c|c}
     
\begin{tabular}{@{}c|c|ccccc|c@{}}
\midrule
Positive & Negative&\multicolumn{6}{c}{Outliers} \\
\hspace{.25em}Class &\hspace{.25em}Class & Plane & Car  & Bird & Cat  & Deer & avg\\
\midrule

\multirow{8}{*}{Plane} &None                           & -     & 0.78 & 0.58 & 0.62 & 0.61 & 0.648                     \\
                       & Dog    5                  & -     & 0.83 & 0.87 & 0.95 & 0.91 & 0.890                \\
                       & Frog   6                  & -     & 0.83 & 0.86 & 0.94 & 0.90 & 0.883                \\
                       & Horse 7                   & -     & 0.82 & 0.86 & 0.94 & 0.91 & 0.883                \\
                       & Ship   8                  & -     & 0.83 & 0.86 & 0.92 & 0.90 & 0.878                \\
                       & Truck 9                   & -     & 0.84 & 0.86 & 0.95 & 0.91 & 0.890                \\
                       & Comb. (5-9)               & -     & 0.83 & 0.85 & 0.93 & 0.91 & 0.880                \\
                       & Sup. (0-4)                & -     & 0.81 & 0.88 & 0.94 & 0.92 & 0.888                \\
                       \midrule
\multirow{8}{*}{Car}   & None                      & 0.32  & -    & 0.34 & 0.33 & 0.33 & 0.330                \\
                       & Dog    5                  & 0.67  & -    & 0.89 & 0.93 & 0.9  & 0.848                \\
                       & Frog   6                  & 0.58  & -    & 0.9  & 0.9  & 0.91 & 0.823                \\
                       & Horse 7                   & 0.66  & -    & 0.88 & 0.9  & 0.92 & 0.840                \\
                       & Ship   8                  & 0.83  & -    & 0.59 & 0.51 & 0.5  & 0.608                \\
                       & Truck 9                   & 0.51  & -    & 0.44 & 0.49 & 0.44 & 0.470                \\
                       & Comb. (5-9)               & 0.81  & -    & 0.92 & 0.92 & 0.94 & 0.898                \\
                       & Sup. (0-4)                & 0.89  & -    & 0.93 & 0.90 & 0.95 & 0.918                \\
                       \midrule
\multirow{8}{*}{Bird}  & None                      & 0.52  & 0.78 & -    & 0.54 & 0.52 & 0.590                \\
                       & Dog    5                  & 0.59  & 0.75 & -    & 0.71 & 0.49 & 0.635                \\
                       & Frog   6                  & 0.53  & 0.78 & -    & 0.69 & 0.54 & 0.635                \\
                       & Horse 7                   & 0.63  & 0.80 & -    & 0.66 & 0.57 & 0.665                \\
                       & Ship   8                  & 0.86  & 0.89 & -    & 0.61 & 0.45 & 0.703                \\
                       & Truck 9                   & 0.76  & 0.94 & -    & 0.64 & 0.72 & 0.765                \\
                       & Comb. (5-9)               & 0.78  & 0.90 & -    & 0.62 & 0.49 & 0.698                \\
                       & Sup. (0-4)                & 0.82  & 0.94 & -    & 0.60 & 0.48 & 0.710                \\
                       \midrule
\multirow{8}{*}{Cat}   & None                      & 0.55  & 0.76 & 0.50 & -    & 0.50 & 0.578                \\
                       & Dog    5                  & 0.54  & 0.73 & 0.52 & -    & 0.52 & 0.578                \\
                       & Frog   6                  & 0.56  & 0.72 & 0.60 & -    & 0.62 & 0.625                \\
                       & Horse 7                   & 0.70  & 0.75 & 0.59 & -    & 0.68 & 0.680                \\
                       & Ship   8                  & 0.91  & 0.89 & 0.53 & -    & 0.46 & 0.678                \\
                       & Truck 9                   & 0.82  & 0.94 & 0.50 & -    & 0.48 & 0.685                \\
                       & Comb. (5-9)               & 0.89  & 0.91 & 0.55 & -    & 0.52 & 0.718                \\
                       & Sup. (0-4)                & 0.93  & 0.93 & 0.58 & -    & 0.54 & 0.745                \\
                       \midrule
\multirow{8}{*}{Deer}  & None                      & 0.56  & 0.80 & 0.52 & 0.54 & -    & 0.605                \\
                       & Dog    5                  & 0.72  & 0.85 & 0.63 & 0.80 & -    & 0.750                \\
                       & Frog   6                  & 0.66  & 0.86 & 0.60 & 0.75 & -    & 0.718                \\
                       & Horse 7                   & 0.71  & 0.58 & 0.58 & 0.71 & -    & 0.645                \\
                       & Ship   8                  & 0.93  & 0.94 & 0.63 & 0.72 & -    & 0.805                \\
                       & Truck 9                   & 0.84  & 0.97 & 0.62 & 0.72 & -    & 0.773                \\
                       & Comb. (5-9)               & 0.87  & 0.95 & 0.61 & 0.73 & -    & 0.790                \\
                       & Sup. (0-4)                & 0.93  & 0.97 & 0.62 & 0.72 & -    & 0.810               \\
\midrule
\end{tabular}
     
     &
     \begin{tabular}{@{}c|c|ccccc|c@{}}
\midrule
Positive & Negative&\multicolumn{6}{c}{Outliers} \\
\hspace{.25em}Class &\hspace{.25em}Class & Dog & Frog  & Horse & Ship  & Truck & avg\\
\midrule
\multirow{8}{*}{Dog}   &None                           & -    & 0.69 & 0.66  & 0.57 & 0.77  & 0.673                     \\
                       & Plane 0                   & -    & 0.53 & 0.66  & 0.92 & 0.89  & 0.750                \\
                       & Car 1                     & -    & 0.56 & 0.68  & 0.95 & 0.91  & 0.775                \\
                       & Bird 2                    & -    & 0.63 & 0.63  & 0.77 & 0.78  & 0.703                \\
                       & Cat 3                     & -    & 0.67 & 0.65  & 0.66 & 0.81  & 0.698                \\
                       & Deer 4                    & -    & 0.73 & 0.69  & 0.70 & 0.76  & 0.720                \\
                       & Comb. (0-4)               & -    & 0.58 & 0.67  & 0.95 & 0.94  & 0.785                \\
                       & Sup. (5-9)                & -    & 0.56 & 0.73  & 0.95 & 0.95  & 0.798                \\
\midrule
\multirow{8}{*}{Frog}  & None                      & 0.40 & -    & 0.53  & 0.49 & 0.67  & 0.523                \\
                       & Plane 0                   & 0.73 & -    & 0.81  & 0.96 & 0.93  & 0.858                \\
                       & Car 1                     & 0.74 & -    & 0.83  & 0.96 & 0.97  & 0.875                \\
                       & Bird 2                    & 0.80 & -    & 0.84  & 0.91 & 0.85  & 0.850                \\
                       & Cat 3                     & 0.84 & -    & 0.80  & 0.87 & 0.86  & 0.843                \\
                       & Deer 4                    & 0.75 & -    & 0.86  & 0.88 & 0.87  & 0.840                \\
                       & Comb. (0-4)               & 0.75 & -    & 0.84  & 0.97 & 0.95  & 0.877                \\
                       & Sup. (5-9)                & 0.82 & -    & 0.88  & 0.97 & 0.96  & 0.907                \\
\midrule
\multirow{8}{*}{Horse} & None                      & 0.41 & 0.58 & -     & 0.46 & 0.66  & 0.528                \\
                       & Plane 0                   & 0.55 & 0.50 & -     & 0.93 & 0.83  & 0.703                \\
                       & Car 1                     & 0.56 & 0.58 & -     & 0.90 & 0.93  & 0.743                \\
                       & Bird 2                    & 0.62 & 0.73 & -     & 0.80 & 0.71  & 0.715                \\
                       & Cat 3                     & 0.77 & 0.76 & -     & 0.65 & 0.66  & 0.710                \\
                       & Deer 4                    & 0.62 & 0.83 & -     & 0.57 & 0.60  & 0.655                \\
                       & Comb. (0-4)               & 0.51 & 0.57 & -     & 0.89 & 0.88  & 0.713                \\
                       & Sup. (5-9)                & 0.59 & 0.66 & -     & 0.95 & 0.92  & 0.780                \\
\midrule
\multirow{8}{*}{Ship}  & None                      & 0.62 & 0.74 & 0.73  & -    & 0.77  & 0.715                \\
                       & Plane 0                   & 0.75 & 0.75 & 0.82  & -    & 0.74  & 0.765                \\
                       & Car 1                     & 0.84 & 0.89 & 0.90  & -    & 0.88  & 0.878                \\
                       & Bird 2                    & 0.94 & 0.96 & 0.94  & -    & 0.78  & 0.905                \\
                       & Cat 3                     & 0.95 & 0.95 & 0.93  & -    & 0.8   & 0.908                \\
                       & Deer 4                    & 0.92 & 0.96 & 0.94  & -    & 0.78  & 0.900                \\
                       & Comb. (0-4)               & 0.95 & 0.97 & 0.96  & -    & 0.83  & 0.928                \\
                       & Sup. (5-9)                & 0.94 & 0.96 & 0.96  & -    & 0.88  & 0.935                \\
\midrule
\multirow{8}{*}{Truck} & None                      & 0.35 & 0.53 & 0.46  & 0.30 & -     & 0.41                 \\
                       & Plane 0                   & 0.61 & 0.52 & 0.58  & 0.80 & -     & 0.628                \\
                       & Car 1                     & 0.51 & 0.57 & 0.53  & 0.47 & -     & 0.520                \\
                       & Bird 2                    & 0.92 & 0.90 & 0.84  & 0.73 & -     & 0.848                \\
                       & Cat 3                     & 0.95 & 0.90 & 0.82  & 0.61 & -     & 0.820                \\
                       & Deer 4                    & 0.91 & 0.92 & 0.86  & 0.63 & -     & 0.830                \\
                       & Comb. (0-4)               & 0.93 & 0.91 & 0.83  & 0.76 & -     & 0.858                \\
                       & Sup. (5-9)                & 0.94 & 0.95 & 0.90  & 0.78 & -     & 0.893\\
\midrule
\end{tabular}
\end{tabular}

\begin{tabular}{c|c}
     
     \begin{tabular}{@{}c|c|ccccc|c@{}}
\midrule
Positive & Negative&\multicolumn{6}{c}{Outliers} \\
\hspace{.25em}Class &\hspace{.25em}Class & Plane & Car  & Bird & Cat  & Deer & avg\\
\midrule
\multirow{8}{*}{Plane} &None&	-&	0.78&	0.58&	0.62&	0.61&	0.648\\
& Dog 5                     & -     & 0.84 & 0.88 & 0.97 & 0.93 & 0.905                \\
                       & Frog 6                    & -     & 0.84 & 0.86 & 0.95 & 0.93 & 0.895                \\
                       & Horse 7                   & -     & 0.83 & 0.85 & 0.95 & 0.92 & 0.888                \\
                       & Ship 8                    & -     & 0.80 & 0.57 & 0.74 & 0.56 & 0.668                \\
                       & Truck 9                   & -     & 0.92 & 0.71 & 0.87 & 0.74 & 0.810                \\
                       & Comb. (0-4)               & -     & 0.90 & 0.85 & 0.96 & 0.94 & 0.913                \\
                       & Sup. (5-9)                & -     & 0.93 & 0.90 & 0.96 & 0.95 & 0.935                \\
                       \midrule
\multirow{8}{*}{Car}   & None                      & 0.32  & -    & 0.34 & 0.33 & 0.33 & 0.330                \\
                       & Dog 5                     & 0.67  & -    & 0.94 & 0.97 & 0.95 & 0.883                \\
                       & Frog 6                    & 0.58  & -    & 0.93 & 0.96 & 0.96 & 0.858                \\
                       & Horse 7                   & 0.69  & -    & 0.95 & 0.97 & 0.97 & 0.895                \\
                       & Ship 8                    & 0.90  & -    & 0.78 & 0.79 & 0.76 & 0.808                \\
                       & Truck 9                   & 0.59  & -    & 0.77 & 0.82 & 0.73 & 0.728                \\
                       & Comb. (0-4)               & 0.90  & -    & 0.97 & 0.97 & 0.98 & 0.955                \\
                       & Sup. (5-9)                & 0.95  & -    & 0.98 & 0.98 & 0.98 & 0.973                \\
                       \midrule
\multirow{8}{*}{Bird}  & None                      & 0.52  & 0.78 & -    & 0.54 & 0.52 & 0.590                \\
                       & Dog 5                     & 0.56  & 0.78 & -    & 0.81 & 0.55 & 0.675                \\
                       & Frog 6                    & 0.53  & 0.80 & -    & 0.71 & 0.56 & 0.650                \\
                       & Horse 7                   & 0.63  & 0.81 & -    & 0.72 & 0.59 & 0.688                \\
                       & Ship 8                    & 0.86  & 0.93 & -    & 0.64 & 0.47 & 0.725                \\
                       & Truck 9                   & 0.74  & 0.95 & -    & 0.68 & 0.48 & 0.713                \\
                       & Comb. (0-4)               & 0.82  & 0.95 & -    & 0.76 & 0.60 & 0.783                \\
                       & Sup. (5-9)                & 0.89  & 0.97 & -    & 0.70 & 0.56 & 0.773                \\
                       \midrule
\multirow{8}{*}{Cat}   & None                      & 0.55  & 0.76 & 0.50 & -    & 0.50 & 0.578                \\
                       & Dog 5                     & 0.50  & 0.71 & 0.51 & -    & 0.46 & 0.545                \\
                       & Frog 6                    & 0.52  & 0.76 & 0.58 & -    & 0.64 & 0.625                \\
                       & Horse 7                   & 0.65  & 0.79 & 0.56 & -    & 0.64 & 0.660                \\
                       & Ship 8                    & 0.93  & 0.92 & 0.56 & -    & 0.48 & 0.723                \\
                       & Truck 9                   & 0.81  & 0.96 & 0.52 & -    & 0.48 & 0.693                \\
                       & Comb. (5-9)               & 0.88  & 0.95 & 0.62 & -    & 0.68 & 0.783                \\
                       & Sup. (5-9)                & 0.95  & 0.97 & 0.75 & -    & 0.76 & 0.860                \\
                       \midrule
\multirow{8}{*}{Deer}  & None                      & 0.56  & 0.80 & 0.52 & 0.54 & -    & 0.605                \\
                       & Dog 5                     & 0.67  & 0.86 & 0.71 & 0.89 & -    & 0.783                \\
                       & Frog 6                    & 0.68  & 0.87 & 0.62 & 0.79 & -    & 0.740                \\
                       & Horse 7                   & 0.70  & 0.84 & 0.61 & 0.72 & -    & 0.718                \\
                       & Ship 8                    & 0.94  & 0.95 & 0.63 & 0.73 & -    & 0.813                \\
                       & Truck 9                   & 0.84  & 0.97 & 0.61 & 0.76 & -    & 0.795                \\
                       & Comb. (0-4)               & 0.90  & 0.97 & 0.66 & 0.80 & -    & 0.833                \\
                       & Sup. (5-9)                & 0.97  & 0.98 & 0.76 & 0.83 & -    & 0.885               
\\
\midrule
\end{tabular}

     & 
     
     \begin{tabular}{@{}c|c|ccccc|c@{}}
\midrule
Positive & Negative&\multicolumn{6}{c}{Outliers} \\
\hspace{.25em}Class &\hspace{.25em}Class & Dog & Frog  & Horse & Ship  & Truck & avg\\
\midrule
\multirow{8}{*}{Dog}   &None                       & -    & 0.69 & 0.66  & 0.57 & 0.77  & 0.673                      \\
                       & Plane 0                   & -    & 0.62 & 0.68  & 0.98 & 0.94  & 0.805                \\
                       & Car 1                     & -    & 0.67 & 0.7   & 0.96 & 0.97  & 0.825                \\
                       & Bird 2                    & -    & 0.73 & 0.68  & 0.94 & 0.86  & 0.803               \\
                       & Cat 3                     & -    & 0.65 & 0.68  & 0.8  & 0.85  & 0.745                \\
                       & Deer 4                    & -    & 0.81 & 0.74  & 0.89 & 0.81  & 0.8125               \\
                       & Comb. (0-4)               & -    & 0.80 & 0.76  & 0.97 & 0.96  & 0.874                \\
                       & Sup. (5-9)                & -    & 0.90 & 0.85  & 0.97 & 0.98  & 0.925                \\
                       \midrule
\multirow{8}{*}{Frog}  & None                      & 0.40 & -    & 0.53  & 0.49 & 0.67  & 0.523                \\
                       & Plane 0                   & 0.7  & -    & 0.58  & 0.98 & 0.96  & 0.841                \\
                       & Car 1                     & 0.7  & -    & 0.83  & 0.98 & 0.98  & 0.930                \\
                       & Bird 2                    & 0.86 & -    & 0.91  & 0.96 & 0.93  & 0.933                \\
                       & Cat 3                     & 0.88 & -    & 0.87  & 0.94 & 0.92  & 0.912                \\
                       & Deer 4                    & 0.81 & -    & 0.92  & 0.95 & 0.92  & 0.929                \\
                       & Comb. (0-4)               & 0.91 & -    & 0.95  & 0.98 & 0.98  & 0.956                \\
                       & Sup. (5-9)                & 0.94 & -    & 0.97  & 0.98 & 0.98  & 0.968                \\
                       \midrule
\multirow{8}{*}{Horse} & None                      & 0.41 & 0.58 & -     & 0.46 & 0.66  & .0531                \\
                       & Plane 0                   & 0.53 & 0.59 & -     & 0.96 & 0.87  & 0.806                \\
                       & Car 1                     & 0.57 & 0.67 & -     & 0.96 & 0.95  & 0.860                \\
                       & Bird 2                    & 0.33 & 0.75 & -     & 0.93 & 0.79  & 0.823                \\
                       & Cat 3                     & 0.78 & 0.81 & -     & 0.9  & 0.77  & 0.826                \\
                       & Deer 4                    & 0.67 & 0.85 & -     & 0.87 & 0.68  & 0.800                \\
                       & Comb. (0-4)               & 0.75 & 0.91 & -     & 0.97 & 0.93  & 0.891                \\
                       & Sup. (5-9)                & 0.82 & 0.96 & -     & 0.98 & 0.96  & 0.930                \\
                       \midrule
\multirow{8}{*}{Ship}  & None                      & 062  & 0.74 & 0.73  & -    & 0.77  & 0.717                \\
                       & Plane 0                   & 0.84 & 0.89 & 0.93  & -    & 0.85  & 0.890                \\
                       & Car 1                     & 0.86 & 0.91 & 0.93  & -    & 0.92  & 0.920                \\
                       & Bird 2                    & 0.96 & 0.97 & 0.97  & -    & 0.85  & 0.930                \\
                       & Cat 3                     & 0.97 & 0.98 & 0.97  & -    & 0.85  & 0.933                \\
                       & Deer 4                    & 0.95 & 0.97 & 0.97  & -    & 0.84  & 0.927                \\
                       & Comb. (0-4)               & 0.97 & 0.98 & 0.98  & -    & 0.90  & 0.956                \\
                       & Sup. (5-9)                & 0.97 & 0.98 & 0.98  & -    & 0.94  & 0.968                \\
                       \midrule
\multirow{8}{*}{Truck} & None                      & 0.35 & 0.53 & 0.46  & 0.30 & -     & 0.412                \\
                       & Plane 0                   & 0.69 & 0.70 & 0.67  & 0.87 & -     & 0.733                \\
                       & Car 1                     & 0.54 & 0.61 & 0.53  & 0.61 & -     & 0.573                \\
                       & Bird 2                    & 0.93 & 0.89 & 0.88  & 0.73 & -     & 0.858                \\
                       & Cat 3                     & 0.96 & 0.91 & 0.88  & 0.63 & -     & 0.845                \\
                       & Deer 4                    & 0.90 & 0.88 & 0.91  & 0.67 & -     & 0.840                \\
                       & Comb. (0-4)               & 0.97 & 0.96 & 0.91  & 0.82 & -     & 0.914                \\
                       & Sup. (5-9)                & 0.98 & 0.98 & 0.96  & 0.89 & -     & 0.953  
\\
\midrule
\end{tabular}
\end{tabular}

\newpage
\section{Detailed Results for all 10 tasks}


\vspace{.2cm}
The following graphs are detailed results for some experiments in various settings described in section 5.1 and 5.2. Each curve represents a trade-off between accuracy on anomalies and on normal data for each dataset. The two left panes are an upper-bound  \textit{supervised} setting where the negative dataset is the same as outliers. The top pane shows accuracies on tasks 0 to 4 and the bottom shows accuracies on tasks 5 to 9. Note that as the threshold value $\alpha$ increases the model favors accepting anomalies over misclassifying normal examples. In almost all cases, we observe that LIS-AE gives a significant margin compared to normal AE.
\\
\begin{tabular}{ll}
     \\
     \\\\
     \includegraphics[trim={2.5cm 2.5cm 2.5cm 2.5cm}, width=.5\linewidth,valign=m]{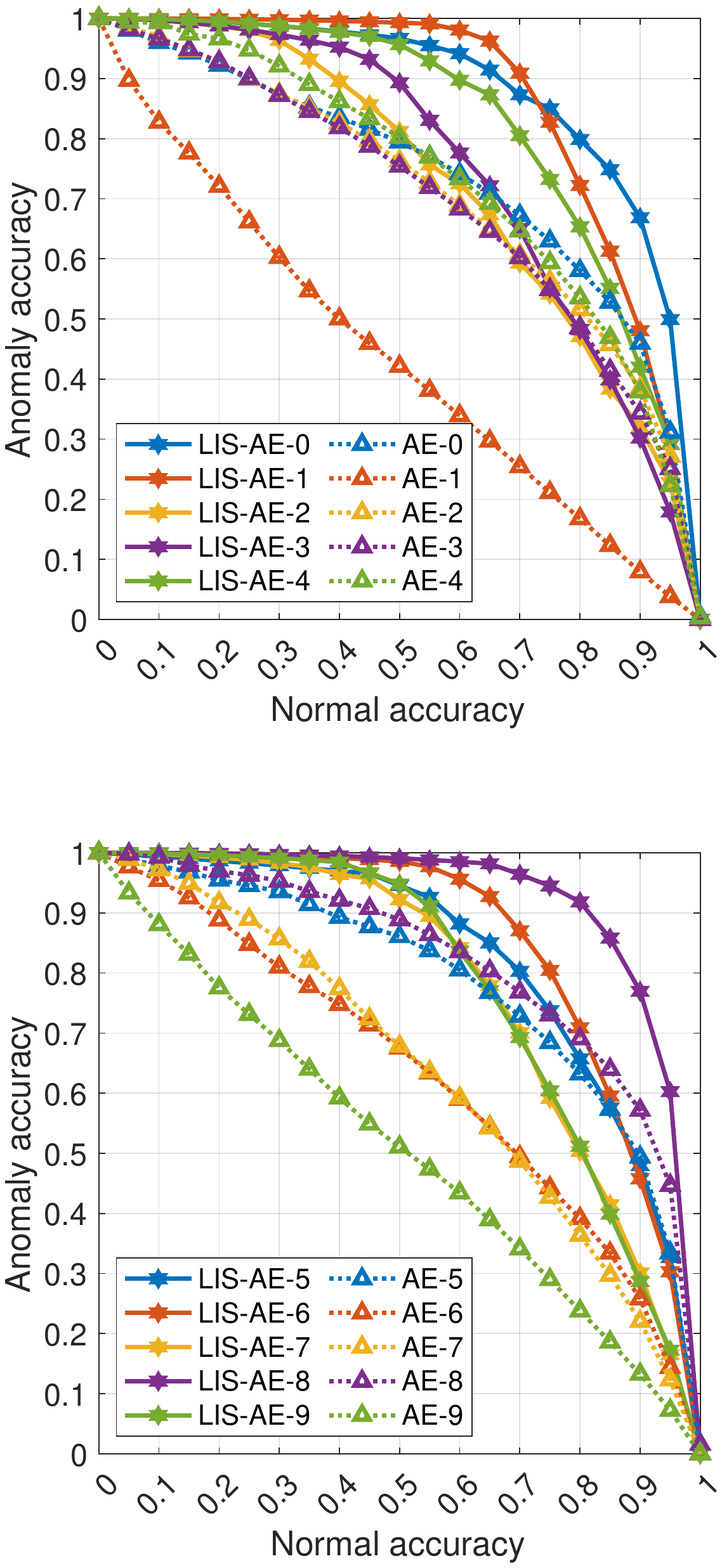}&
     \includegraphics[trim={2.5cm 2.5cm 2.5cm 2.5cm}, width=.5\linewidth,valign=m]{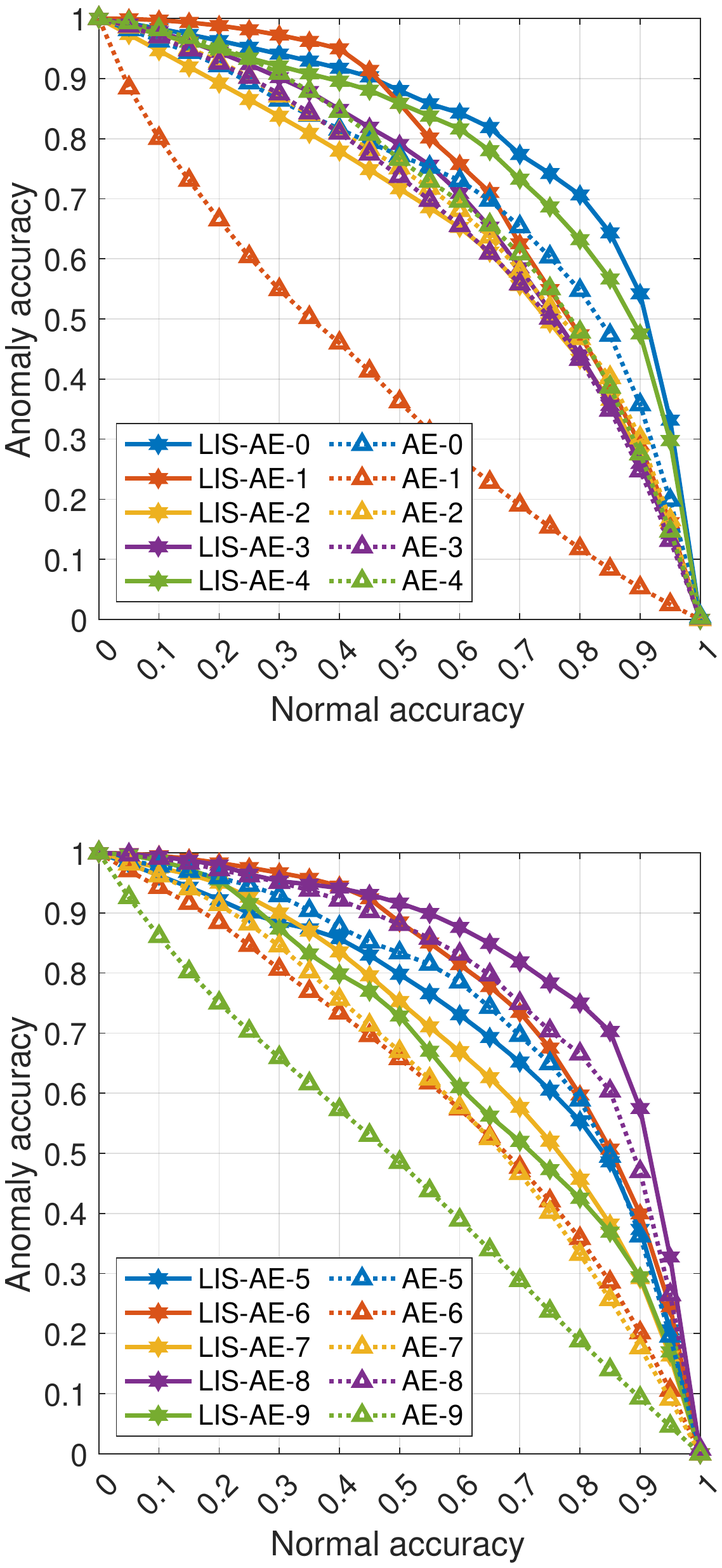}\\\\\\\\
     \multicolumn{2}{c}{Standard LIS-AE trained on CIFAR-10 classes. Left, outliers as negative dataset (Supervised). Right, SVHN as negative dataset.}
\end{tabular}\\

\begin{tabular}{ll}
     \\
     \\\\
     \includegraphics[trim={2.5cm 2.5cm 2.5cm 2.5cm}, width=.5\linewidth,valign=m]{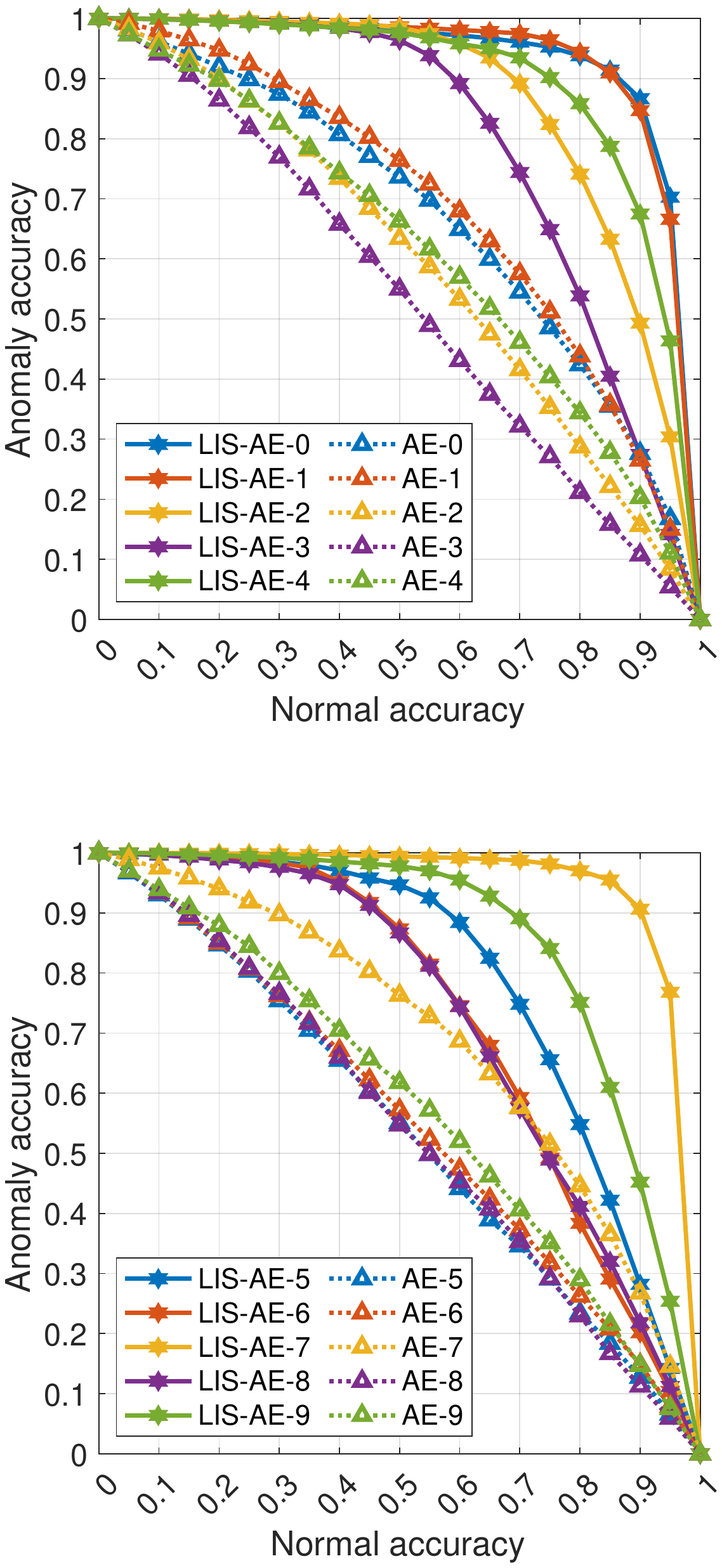}&
     \includegraphics[trim={2.5cm 2.5cm 2.5cm 2.5cm}, width=.5\linewidth,valign=m]{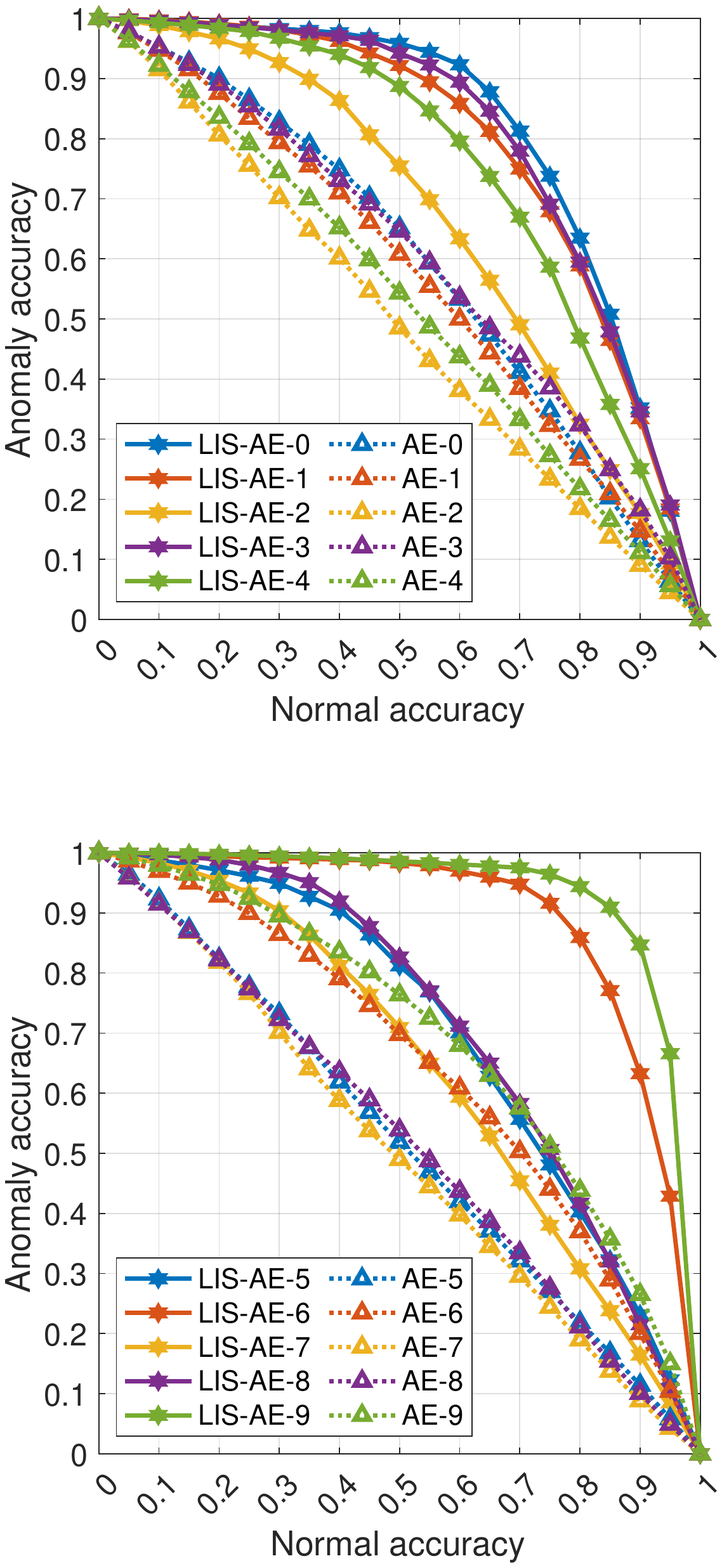}\\\\\\\\
     \multicolumn{2}{c}{Results of LinSep-LIS-AE variant on SVHN. Left, outliers as negative dataset (Supervised). Right, unsupervised.}
\end{tabular}\\

\begin{tabular}{ll}
     \\
     \\\\
     \includegraphics[trim={2.5cm 2.5cm 2.5cm 2.5cm}, width=.5\linewidth,valign=m]{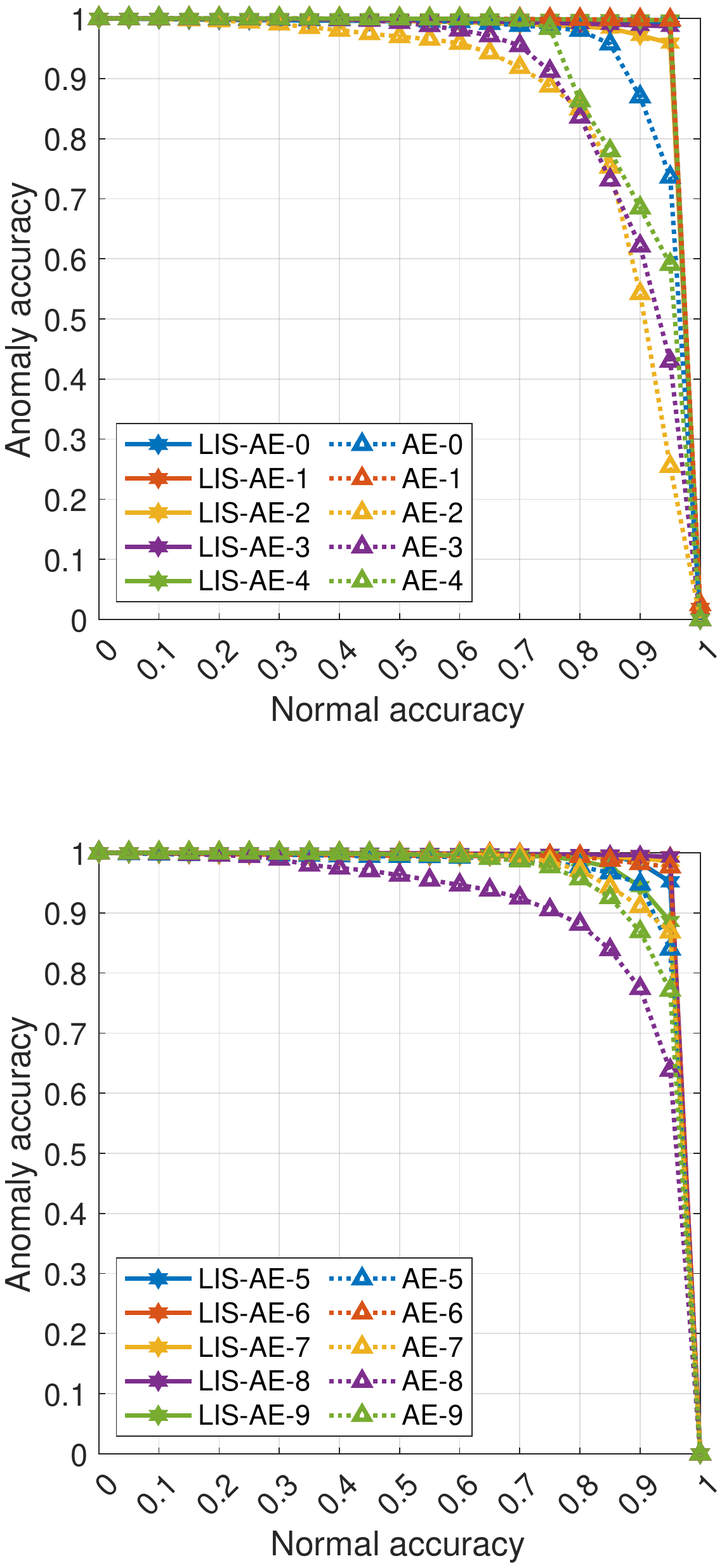}&
     \includegraphics[trim={2.5cm 2.5cm 2.5cm 2.5cm}, width=.5\linewidth,valign=m]{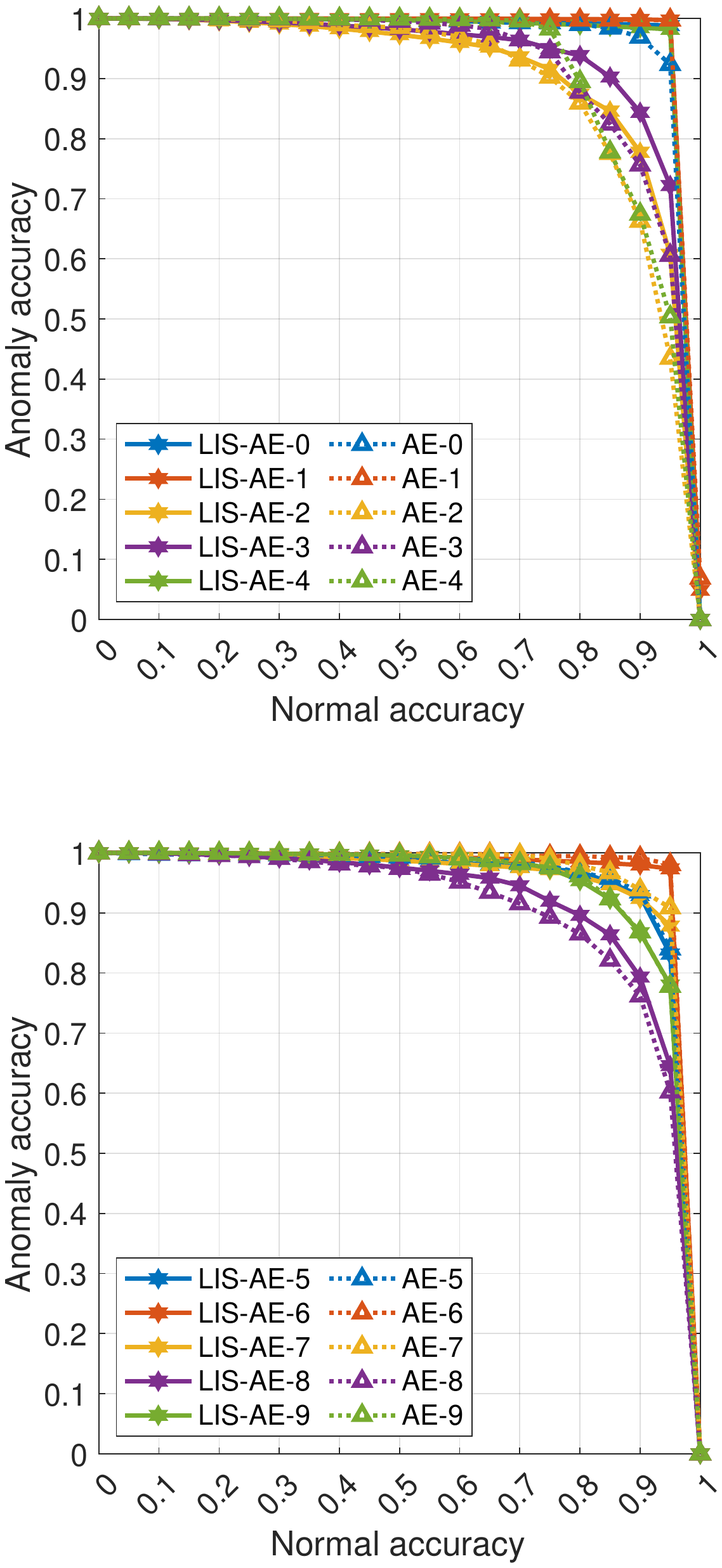}\\\\\\\\
     \multicolumn{2}{c}{MNIST classes as Positive datasets. Left, outliers as negative dataset (Supervised). Right, Omniglot as negative dataset.}
\end{tabular}

\begin{tabular}{ll}
     \\
     \\\\
     \includegraphics[trim={2.5cm 2.5cm 2.5cm 2.5cm}, width=.5\linewidth,valign=m]{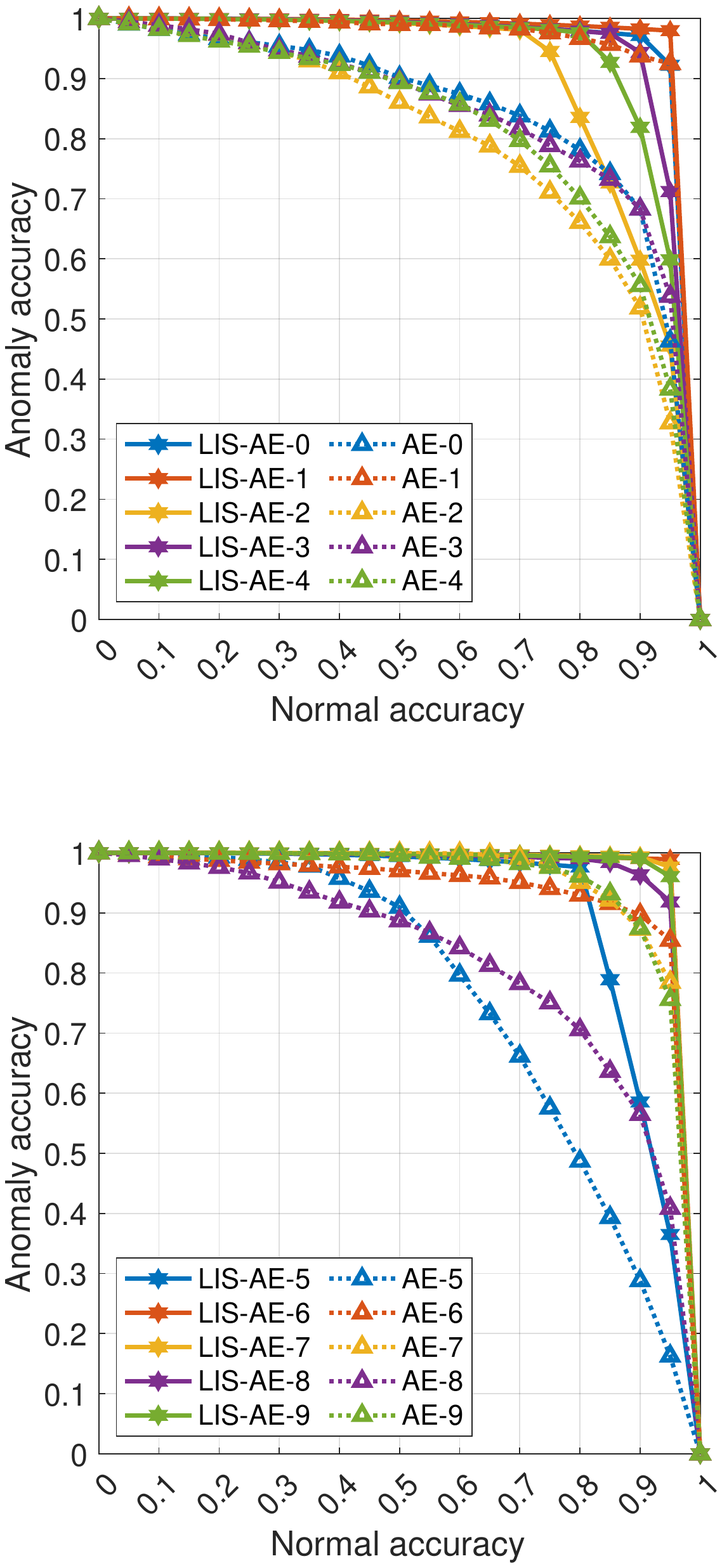}&
     \includegraphics[trim={2.5cm 2.5cm 2.5cm 2.5cm}, width=.5\linewidth,valign=m]{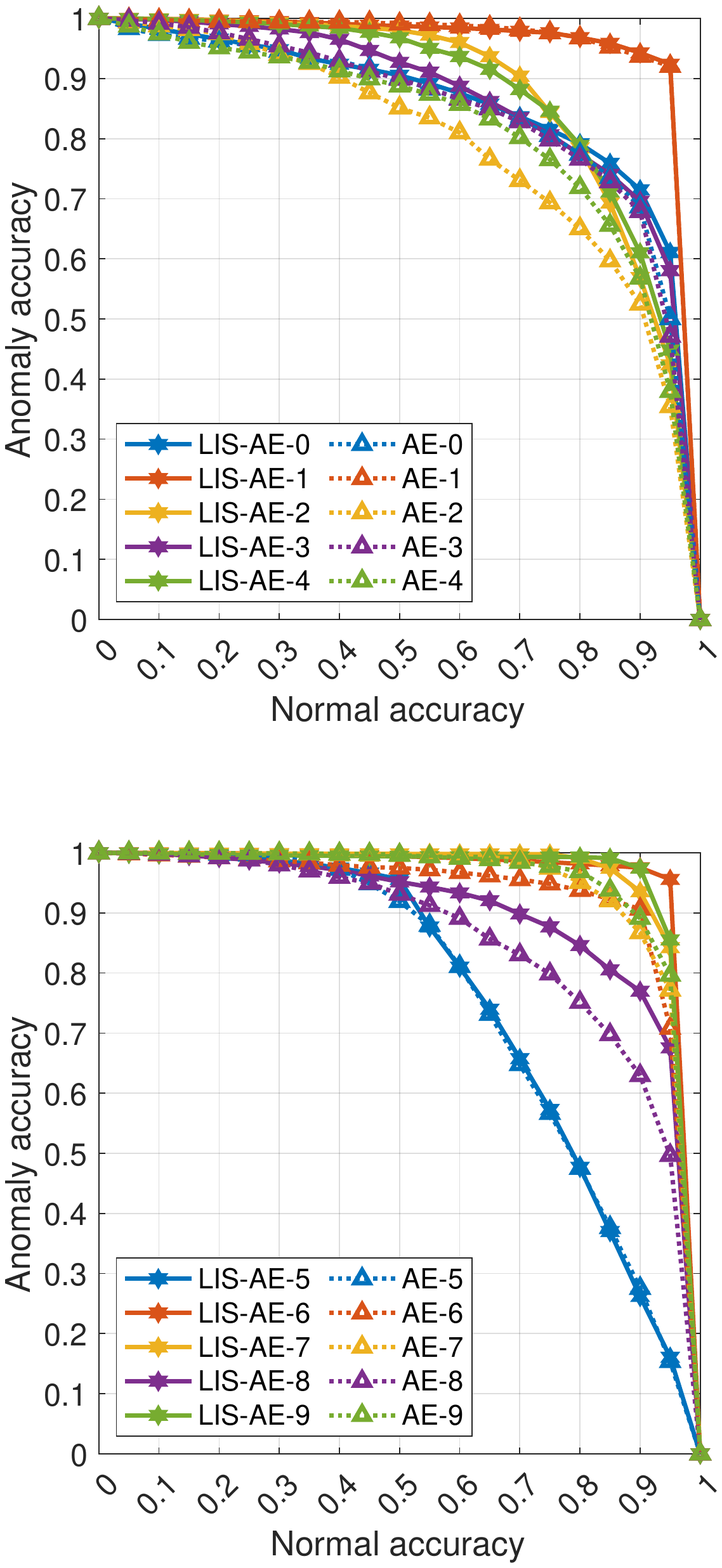}\\\\\\\\
     \multicolumn{2}{c}{Fashion-MNIST classes as Positive datasets. Left, outliers as negative dataset (Supervised). Right, Omniglot as negative dataset.}
\end{tabular}
\\

\end{document}